\documentclass[10pt,twocolumn,letterpaper]{article}

\usepackage{iccv}
\usepackage{times}
\usepackage{epsfig}
\usepackage{graphicx}
\usepackage{amsmath}
\usepackage{amssymb}
\usepackage{amsmath,algorithm,algpseudocode,amsfonts}
\usepackage{epstopdf}
\usepackage{subcaption}
\usepackage{multirow}
\usepackage{booktabs}

\usepackage[breaklinks=true,bookmarks=false]{hyperref}

\iccvfinalcopy 

\def\iccvPaperID{1291} 

\ificcvfinal\fi

\begin{document}
	
	\title{SCRDet: Towards More Robust Detection for Small, Cluttered \\ and Rotated Objects}
	
	\author{Xue Yang$^{1,2,3,4}$, Jirui Yang$^{2}$, Junchi Yan$^{3,4,}$\thanks{Corresponding author is Junchi Yan. The work is partially supported by National Key Research and Development Program of China (2016YFB1001003), STCSM (18DZ1112300), NSFC (61602176, 61725105, 41801349).} , Yue Zhang$^{1}$, Tengfei Zhang$^{1,2}$ \\ Zhi Guo$^{1}$, Xian Sun$^{1}$, Kun Fu$^{1,2}$\\
		$^{1}$NIST, Institute of Electronics, Chinese Academy of Sciences, Beijing (Suzhou), China.\\
		$^{2}$University of Chinese Academy of Sciences, Beijing, China.\\
		$^{3}$Department of Computer Science and Engineering, Shanghai Jiao Tong University.\\
		$^{4}$MoE Key Lab of Artificial Intelligence, AI Institute, Shanghai Jiao Tong University. \\
		{\tt\small \{yangxue-2019-sjtu, yanjunchi\}@sjtu.edu.cn \quad \{yangjirui16, zhangtengfei16\}@mails.ucas.ac.cn} \\
		{\tt\small \quad zhangyue@aircas.ac.cn \quad \{guozhi, sunxian, fukun\}@mail.ie.ac.cn}
	}
	
	\maketitle
	\ificcvfinal\thispagestyle{empty}\fi
	
	\begin{abstract}
		Object detection has been a building block in computer vision. Though considerable progress has been made, there still exist challenges for objects with small size, arbitrary direction, and dense distribution. Apart from natural images, such issues are especially pronounced for aerial images of great importance. This paper presents a novel multi-category rotation detector for small, cluttered and rotated objects, namely SCRDet. Specifically, a sampling fusion network is devised which fuses multi-layer feature with effective anchor sampling, to improve the sensitivity to small objects. Meanwhile, the supervised pixel attention network and the channel attention network are jointly explored for small and cluttered object detection by suppressing the noise and highlighting the objects feature. For more accurate rotation estimation, the IoU constant factor is added to the smooth L1 loss to address the boundary problem for the rotating bounding box. Extensive experiments on two remote sensing public datasets DOTA, NWPU VHR-10 as well as natural image datasets COCO, VOC2007 and scene text data ICDAR2015 show the state-of-the-art performance of our detector. The code and models will be available at \url{https://github.com/DetectionTeamUCAS}.
	\end{abstract}
	
	\begin{figure*}[!tb]
		\begin{center}
			\includegraphics[width=0.9\linewidth, height=3cm]{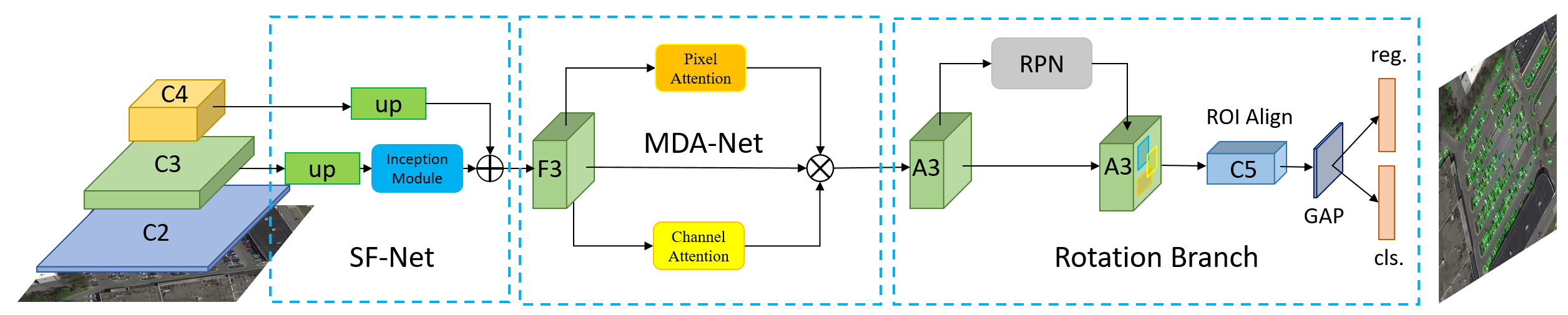}
		\end{center}
		\vspace{-20pt}
		\caption{SCRDet includes SF-Net, MDA-Net against small and cluttered objects and rotation branch for rotated objects.}
		\label{fig:R2CNN++}
	\end{figure*}
	
	\section{Introduction}
	Object detection is one of the fundamental tasks in computer vision and various general-purpose detectors~\cite{girshick2014rich,he2014spatial,girshick2015fast,liu2016ssd,redmon2016you,dai2016r,ren2017faster} have been devised. Promising results have been achieved on a few benchmarks including COCO \cite{lin2014microsoft} and VOC2007 \cite{everingham2010pascal} etc. However, most existing detectors do not pay particular attention to some useful aspects for robust object detection in open environment: small objects, cluttered arrangement and arbitrary orientations.
	
	In real-world problems, due to limitation of camera resolution and other reasons, the objects of interest can be of very small size e.g. for detection of traffic signs, tiny faces under public cameras on the streets. Also, the objects can range in a very dense fashion e.g. goods in shopping malls. Moreover, the objects can no longer be positioned horizontally as in COCO, VOC2007, e.g. for scene text detection whereby the texts can be in any direction and position.
	
	In particular, the above three challenges are pronounced for images in remote sensing, as analyzed as follows:
	
	1) \textbf{Small objects.} Aerial images often contain small objects overwhelmed by complex surrounding scenes;
	
	2) \textbf{Cluttered arrangement.} Objects for detection are often densely arranged, such as vehicles and ships;
	
	3) \textbf{Arbitrary orientations.} Objects in aerial images can appear in various orientations. It is further challenged by the large aspect ratio issue which is common in remote sensing.
	
	In this paper, we mainly discuss our approach in the context of remote sensing, while the approach and the problems are general and we have tested with various datasets beyond aerial images as will be shown in the experiments.
	
	

	Many existing general-purpose detectors such Faster-RCNN \cite{ren2017faster} have been widely employed for aerial object detection. However, the design of such detectors are often based on the implicit assumption that the bounding boxes are basically in horizontal position, which is not the case for aerial images (and other detection tasks e.g. scene text detection). This limitation is further pronounced by the popular non-maximum suppression (NMS) technique as post-processing as it will suppress the detection of densely arranged objects in arbitrary orientation over the horizontal line. Moreover, horizontal region based methods have a coarse resolution on orientation estimation, which is key information to extract for remote sensing.
	
	
	We propose a novel multi-category rotation detector for small, cluttered and rotated objects, called SCRDet which is designated to address the following issues: 1) small object: a sampling fusion network (SF-Net) is devised that incorporates feature fusion and finer anchor sampling; 2) noisy background: a supervised multi-dimensional attention network (MDA-Net) is developed which consists of pixel attention network and channel attention network to suppress the noise and highlight foreground. 3) cluttered and dense objects in arbitrary orientation: an angle sensitive network is devised by introducing an angle related parameter for estimation. Combing these three techniques as a whole, our approach achieves state-of-the-art performance on public datasets including two remote sensing benchmarks DOTA and NWPU VHR-10. The contributions of this paper are:
	
	
	1) For small objects, a tailored feature fusion structure is devised by feature fusion and anchor sampling.
	
	2) For cluttered, small object detection, a supervised multi-dimensional attention network is developed to reduce the adverse impact of background noise.
	
	3) Towards more robust handling of arbitrarily-rotated objects, an improved smooth L1 loss is devised
	by adding the IoU constant factor, which is tailored to solve the boundary problem of the rotating bounding box regression.
	
	4) Perhaps more importantly, in Section \ref{sec:ee} we show that the proposed techniques are general, and can also be applied on natural images and combined with general detection algorithms, which surpass the state-of-the-art method or further improves the existing methods by combination.

	
	\section{Related Work}
	Existing detection methods mainly assume the objects for detection are located along the horizontal line in images. In the seminal work~\cite{girshick2014rich}, a multi-stage R-CNN network for region based detection is presented with a subsequent line of improvements on both accuracy and efficiency, including Fast R-CNN \cite{girshick2015fast}, Faster R-CNN \cite{ren2017faster}, and region-based fully convolutional networks (R-FCN) \cite{dai2016r}. On the other hand, there is also a line of recent works that directly regress the bounding box, e.g. Single-Shot Object Detector (SSD) \cite{liu2016ssd} and You only look once (YOLO) \cite{redmon2016you} leading to improved speed.
	
	As discussed above, there are challenging scenarios regarding with small objects, dense arrangement and arbitrary rotation. However they have not been particularly addressed by the above detectors despite their importance in practice. In particular for aerial images, due to its strategic value to the nation and society, efforts have also been made to develop tailored methods to remote sensing. The R-P-Faster R-CNN framework is developed in \cite{han2017efficient} for small objects. While both deformable convolution layers \cite{dai2017deformable} and R-FCN are combined by \cite{xu2017deformable} to improve detection accuracy. More recently, the authors in \cite{xu2017deformable} adopt top-down and skipped connections to produce a single high-level feature map of a fine resolution, improving the performance of the deformable Faster R-CNN. However such horizontal region based detectors still are confronted with the challenges for the aforementioned bottlenecks in terms of scale, orientation and density, which call for more principled methods beyond the setting for horizontal region detection. On the other hand, there is a thread of works on remote sensing, for detecting objects in arbitrary direction. However, these methods are often tailored to specific object categories, e.g. vehicle \cite{tang2017arbitrary}, ship \cite{yang2018automatic, yang2018position,liu2017rotated,zhang2018toward,liu2018arbitrary}, aircraft \cite{liu2017learning} etc.. Though there are recently a few methods for multi-category rotational region detection models \cite{azimi2018towards, ding2018learning}, while they lack a principled way of handling small size and high density.

	
	Compared with the detection methods for natural images, literature on scene text detection~\cite{jiang2017r2cnn,ma2018arbitrary} often pay more attention to object orientation. While such methods still have difficulty in dealing with aerial image based object detection: one reason is that most text detection methods are restricted to single-category object detection \cite{zhou2017east, Shi2017Detecting,deng2018pixellink}, while there are often many different categories to discern for remote sensing. Another reason is that the objects in aerial images are often more closer to each other than in scene texts, which limits the applicability of segmentation based detection algorithm \cite{deng2018pixellink,zhou2017east} that otherwise work well on scene texts. Moreover, there are often a large number of densely distributed objects that call for efficient detection.

	This paper considers all the above aspects comprehensively, and proposes a principled method for multi-category arbitrary-oriented object detection in aerial images.
	
	\section{The Proposed Method}
	
	We first give an overview of our two-stage method as sketched in Fig. \ref{fig:R2CNN++}. In the first stage, the feature map is expected to contain more feature information and less noise by adding SF-Net and MDA-Net. For positional sensitivity of the angle parameters, this stage still regresses the horizontal box. By the improved five-parameter regression and the rotation nonmaximum-suppression (R-NMS) operation for each proposal in the second stage, we can obtain the final detection results under arbitrary rotations.
	\begin{figure}[!tb]
		\centering
		\begin{subfigure}{.24\textwidth}
			\centering
			\includegraphics[width=.9\linewidth, height=3.8cm]{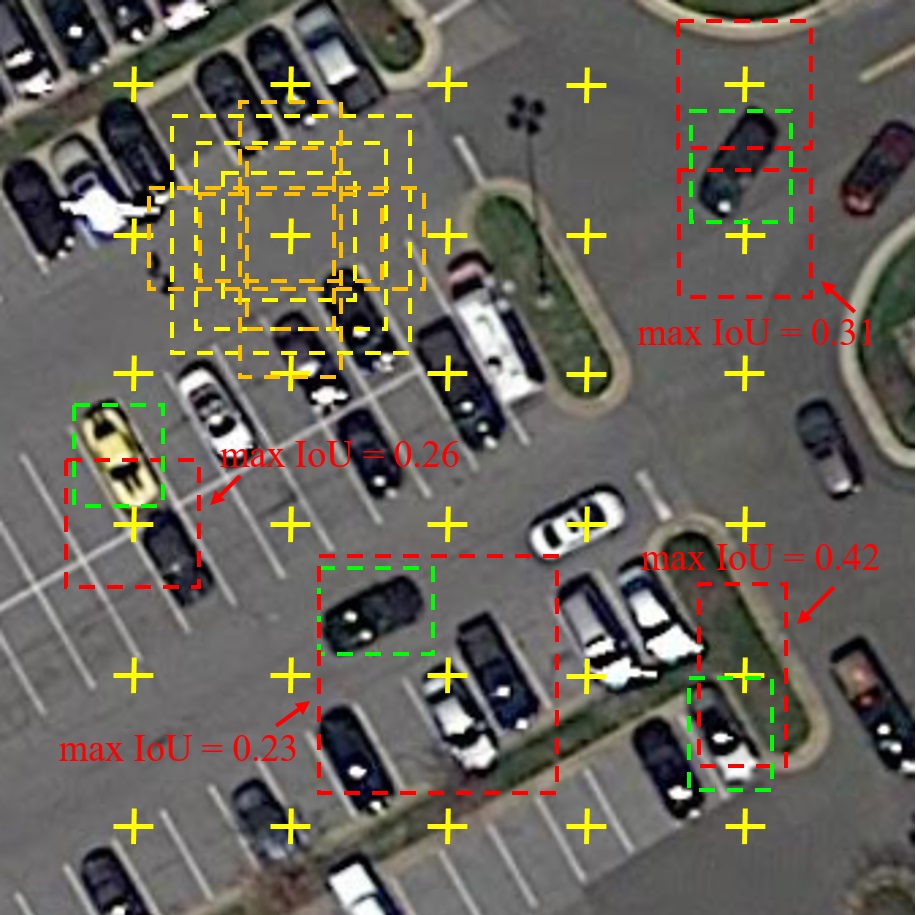}
			\caption{$S_{A}=16$}
			\label{fig:SF-Net-a}
		\end{subfigure}%
		\begin{subfigure}{.24\textwidth}
			\centering
			\includegraphics[width=.9\linewidth, height=3.8cm]{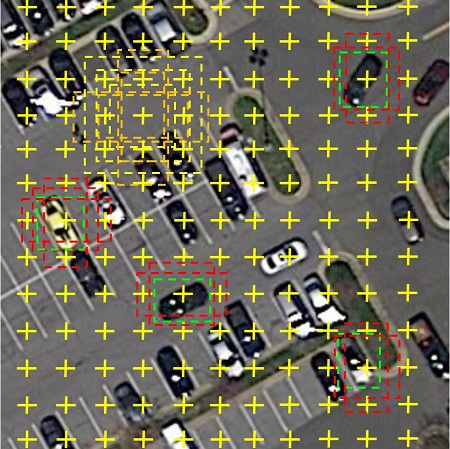}
			\caption{$S_{A}=8$}
			\label{fig:SF-Net-b}
		\end{subfigure}
		\vspace{-10pt}
		\caption{Anchor sampling with different anchor stride $S_{A}$. The orange-yellow bounding box represents the anchor, the green represents ground-truth, and the red box represents the anchor with the largest IoU of ground-truth.}
		\label{fig:SA}
	\end{figure}
	\subsection{Finer Sampling and Feature Fusion Network}\label{subsec:sfn}
	In our analysis, there are two main obstacles in detecting small objects: insufficient object feature information and inadequate anchor samples. The reason is that due to the use of the pooling layer, the small object loses most of its feature information in the deep layers. Meanwhile, larger sampling stride of high-level feature maps tend to skip smaller objects directly, resulting in insufficient sampling.
	
	{\bf Feature fusion.} It is generally regarded that low-level feature map can preserve location information of small object, while high-level feature map can contain higher-level semantic cues. Feature pyramid networks (FPN)~\cite{lin2017feature}, Top-Down Modulation (TDM)~\cite{shrivastava2016beyond}, and Reverse connection with objectness prior networks (RON)~\cite{kong2017ron} are common feature fusion methods that involve the combination of both high and low level feature maps in different forms. 
	
	{\bf Finer sampling.} Insufficient training samples and imbalance can affect the detection performance. By introducing the expected max overlapping (EMO) score, the authors in \cite{zhu2018seeing} calculate the expected max intersection over union (IoU) between anchor and object. They find the smaller stride of the anchor ($S_{A}$) is, the higher EMO score achieves, statistically leading to improved average max IoU of all objects. Fig. \ref{fig:SA} shows the results of small object sampling given stride step 16 and 8, respectively. It can be seen that a smaller $S_{A}$ can sample more high-quality samples well capturing the small objects which is of help for both detector training and inference.
	
	\begin{figure}[!tb]
		\begin{center}
			\includegraphics[width=0.8\linewidth, height=8.5cm]{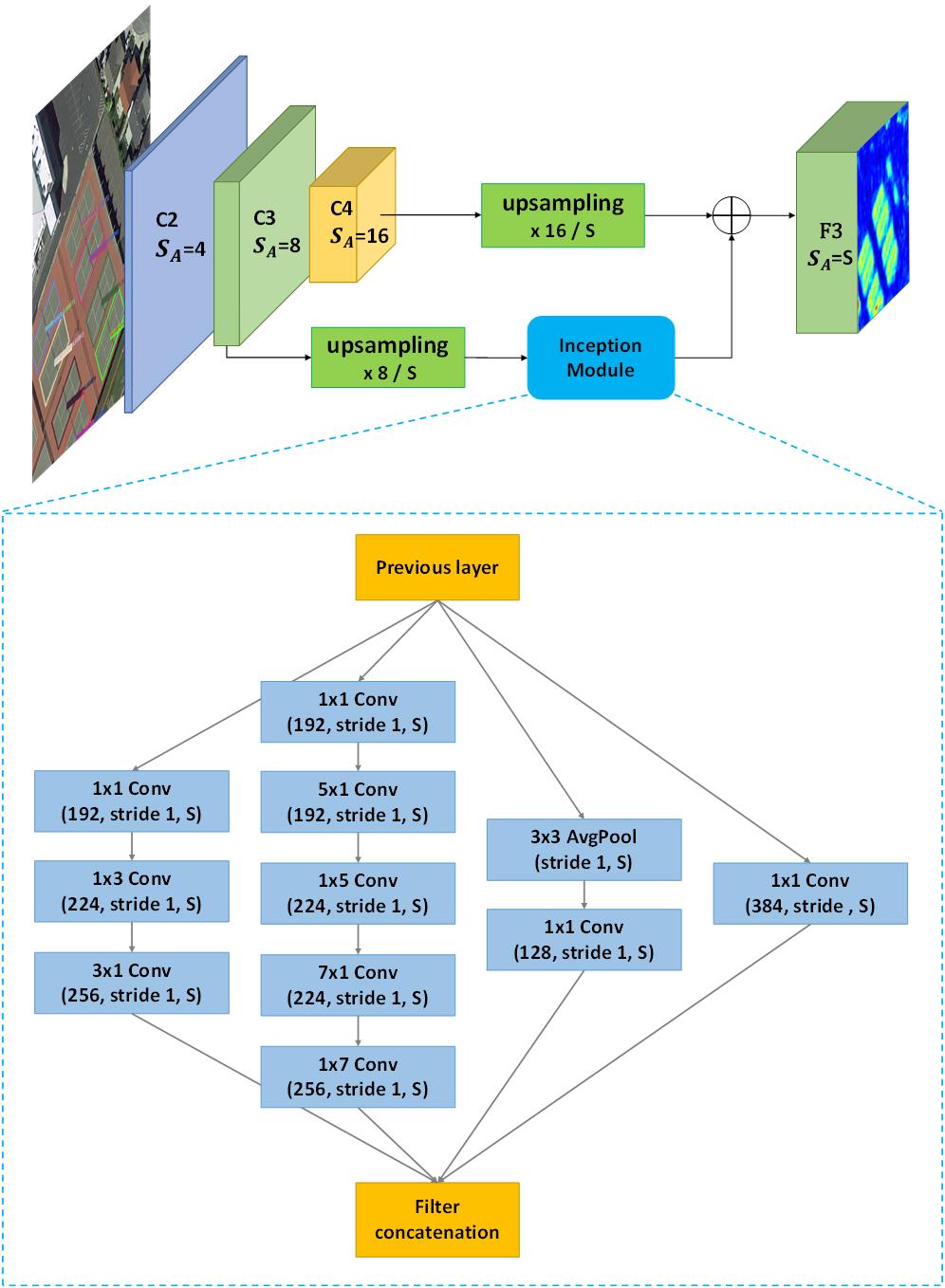}
		\end{center}
		\vspace{-20pt}
		\caption{SF-Net. F3 has a small $S_{A}$, while fully considering the feature fusion and adaptability to different scales.}
		\label{fig:SF-Net}
	\end{figure}
	
	Based on the above analysis, we design the finer sampling and feature fusion network (SF-Net) as shown in Fig.~\ref{fig:SF-Net}. In the anchor based detection framework, the value of $S_{A}$ is equal to the reduction factor of the feature map relative to the original image. In other words, the value of $S_{A}$ can only be an exponential multiple of 2. SF-Net solves this problem by changing the size of the feature map, making the setting of $S_{A}$ more flexible to allow for more adaptive sampling. For the purpose of reducing network parameters, SF-Net only uses C3 and C4 in Resnet ~\cite{he2016deep} for fusion to balance the semantic information and location information while ignoring other less relevant features. In simple terms, the first channel of SF-Net upsamples the C4 so that its $S_{A}=S$, where $S$ is the expected anchor stride. The second channel also upsamples the C3 to the same size. Then, we pass C3 through an inception structure to expand its receptive field and increase semantic information. The inception structure contains a variety of ratio convolution kernels to capture the diversity of object shapes. Finally, a new feature map F3 is obtained by element-wise addition of the two channels. Table \ref{table:SA_TS} shows the detection accuracy and training overhead on DOTA under different $S_{A}$. We find that the optimal $S_{A}$ depends on specific dataset, especially on the size distribution of small objects. In this paper, the value of $S$ is universally set to $6$ for tradeoff between accuracy and speed.
	
	\begin{figure}[!tb]
		\centering
		\begin{subfigure}{.15\textwidth}
			\centering
			\includegraphics[width=.98\linewidth]{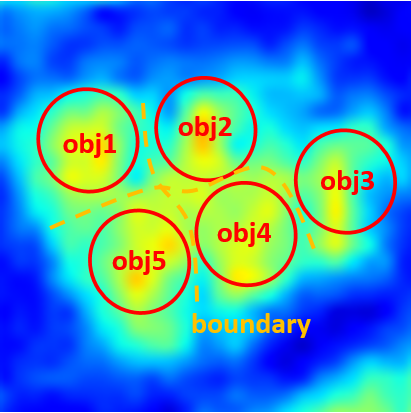}
			\caption{}
			\label{fig:attention_a}
		\end{subfigure}
		\begin{subfigure}{.15\textwidth}
			\centering
			\includegraphics[width=.98\linewidth]{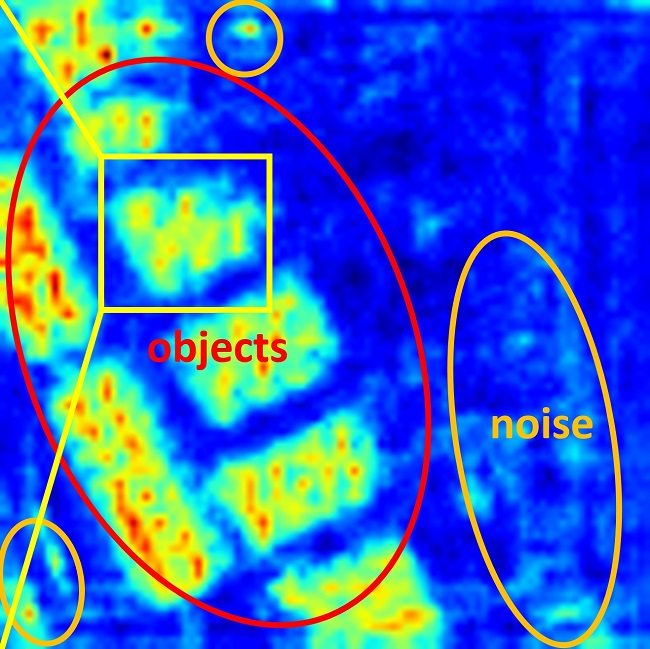}
			\caption{}
			\label{fig:attention_b}
		\end{subfigure}
		\begin{subfigure}{.15\textwidth}
			\centering
			\includegraphics[width=.98\linewidth]{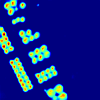}
			\caption{}
			\label{fig:attention_c}
		\end{subfigure} \\
		\begin{subfigure}{.15\textwidth}
			\centering
			\includegraphics[width=.98\linewidth]{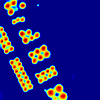}
			\caption{}
			\label{fig:attention_d}
		\end{subfigure}
		\begin{subfigure}{.15\textwidth}
			\centering
			\includegraphics[width=.98\linewidth]{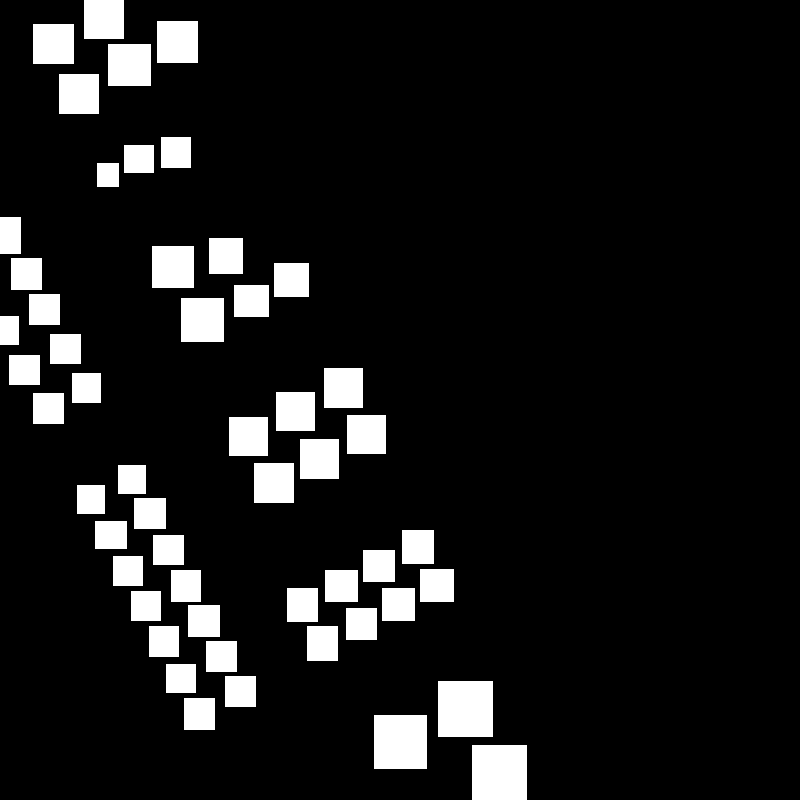}
			\caption{}
			\label{fig:attention_e}
		\end{subfigure}
		\begin{subfigure}{.15\textwidth}
			\centering
			\includegraphics[width=.98\linewidth]{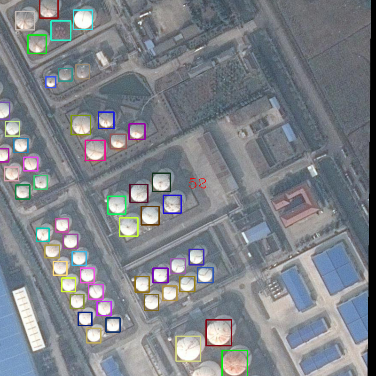}
			\caption{}
			\label{fig:attention_f}
		\end{subfigure}
		\vspace{-10pt}
		\caption{Visualization of the multi-dimensional attention network. (a) Blurred boundaries. (b) Input feature map of attention network. (c) Output feature map of attention network. (d) Saliency map. (e) Binary map. (f) Ground-truth.}
		\label{fig:attention_visualize}
	\end{figure}
	\begin{table}[!tb]
		\begin{center}
			\vspace{-10pt}
			\resizebox{0.48\textwidth}{!}{
				\begin{tabular}{l|rrrrrr}
					\toprule
					anchor stride $S_{A}$ & 6 & 8 & 10 & 12 & 14 & 16 \\
					\hline
					OBB mAP (\%)  & {\bf 67.06} & 66.88 & 65.32 & 63.75 & 63.32 & 63.64 \\
					\hline
					HBB mAP (\%) & {\bf 70.71} & 70.19 & 68.96 & 69.09 & 68.54 & 69.33 \\
					\hline
					Training time (sec.) & 1.18 & 0.99 & 0.76 & 0.46 & 0.39 & {\bf 0.33}  \\		
					\bottomrule
			\end{tabular}}
		\end{center}
		\vspace{-20pt}
		\caption{Accuracy and average training overhead per image with 18K iterations on DOTA under varying stride $S_{A}$.}
		\label{table:SA_TS}
	\end{table}
	\subsection{Multi-Dimensional Attention Network}\label{subsec:mdan}
	Due to the complexity of real-world data such as aerial images, the proposals provided by RPN may introduce a large amount of noise information, as shown in Fig.~\ref{fig:attention_b}. Excessive noise can overwhelm the object information, and the boundaries between the objects will be blurred (see Fig.~\ref{fig:attention_a}), resulting in missed detection and increasing false alarms. Therefore, it is necessary to enhance the object cues and weaken the non-object information. Many attention structures \cite{hu2017squeeze,hu2018relation,wang2017face,Wang2017Non} have been proposed to solve problems of occlusion, noise, and blurring. However, most of the methods are unsupervised, which have difficulty to guide the network to learn specific purposes.
	
\begin{figure}[!tb]	
		\centering
		\includegraphics[width=0.9\linewidth]{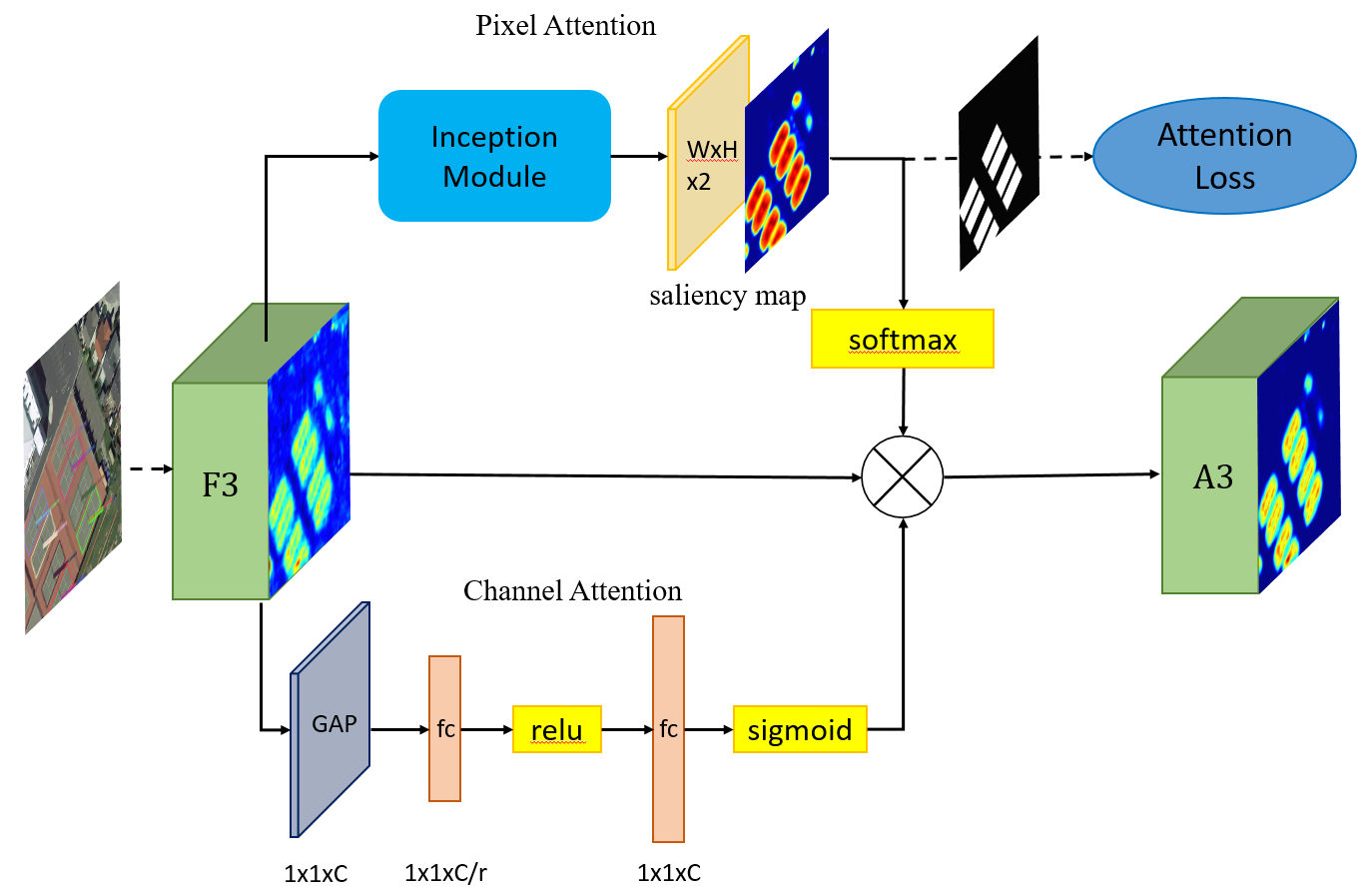}
		\vspace{-10pt}
		\caption{The devised MDA-Net consisting of channel attention network and pixel attention network.}
		\label{fig:attention}
	\end{figure}

	To more effectively capture the objectness of small objects against complex background, we design a supervised multi-dimensional attention leaner (MDA-Net), as shown in Fig. \ref{fig:attention}. Specifically, in the pixel attention network, the feature map F3 passes through an inception structure with different ratio convolution kernels, and then a two-channel saliency map is learned (see Fig.~\ref{fig:attention_d}) through a convolution operation. The saliency map represents the scores of the foreground and background, respectively. Then, Softmax operation is performed on the saliency map and one of the channels is selected to multiply with F3. Finally, a new information feature map A3 is obtained, as shown in Fig.~\ref{fig:attention_c}. It should be noted that the value of the saliency map after the Softmax function is between [0, 1]. In other words, it can reduce the noise and relatively enhance the object information. Since the saliency map is continuous, non-object information will not be eliminated entirely, which is beneficial to retain certain context information and improve robustness. To guide the network to learning this process, we adopt a supervised learning method. Firstly, we can easily get a binary map as a label (as shown in Fig.~\ref{fig:attention_e}) according to ground truth, and then use the cross-entropy loss of the binary map and the saliency map as the attention loss. Besides, we also use SENet~\cite{hu2017squeeze} as the channel attention network for auxiliary, and the value of reduction ratio is 16.
	
	\subsection{Rotation Branch}
	The RPN network provides coarse proposals for the second stage. In order to improve the calculation speed of RPN, we take the highest score of 12,000 regression boxes for NMS operation in the training stage and get 2,000 as proposals. In the test stage, 300 proposals are taken from 10,000 regression boxes by NMS.
	
	In the second stage, we use five parameters ($x,y,w,h,\theta$) to represent arbitrary-oriented rectangle. Ranging in $[-\pi/2,0)$, $\theta$ is defined as the acute angle to the x-axis, and for the other side we denote it as $w$. This definition is consistent with OpenCV. Therefore, IoU computation on axis-aligned bounding box may lead to an inaccurate IoU of the skew interactive bounding box and further ruin the bounding box prediction. An implementation for skew IoU computation \cite{ma2018arbitrary} with thought to triangulation is proposed to deal with this problem. We use rotation nonmaximum-suppression (R-NMS) as a post-processing operation based on skew IoU computation. For the diversity of shapes in the dataset, we set different R-NMS thresholds for different categories. In addition, to make full use of the pre-training weight ResNet, we replace the two fully connected layers fc6 and fc7 with C5 block and global average pooling (GAP). The regression of the rotation bounding box is:
	\begin{equation}
	\begin{aligned}
	t_{x}&=(x-x_{a})/w_{a}, t_{y}=(y-y_{a})/h_{a} \\
	t_{w}&=\log(w/w_{a}), t_{h}=\log(h/h_{a}), t_{\theta}=\theta-\theta_{a}
	\label{eq:regression1}
	\end{aligned}
	\end{equation}
	\begin{equation}
	\begin{aligned}
	t_{x}^{'}&=(x_{}^{'}-x_{a})/w_{a}, t_{y}^{'}=(y_{}^{'}-y_{a})/h_{a} \\
	t_{w}^{'}&=\log(w_{}^{'}/w_{a}), t_{h}^{'}=\log(h_{}^{'}/h_{a}), t_{\theta}^{'}=\theta_{}^{'}-\theta_{a}
	\label{eq:regression2}
	\end{aligned}
	\end{equation}
	where $x,y,w,h,\theta$ denote the box's center coordinates, width, height and angle, respectively. Variables $x, x_{a}, x^{'}$ are for the ground-truth box, anchor box, and predicted box, respectively (likewise for $y,w,h,\theta$).
	
	\begin{table*}[!tb]
		\centering
		\resizebox{0.98\textwidth}{!}{
			\begin{tabular}{|l|c|c|c|c|c|c|c|c|c|c|c|c|c|c|c|c|}
				\toprule
				Method &  PL &  BD &  BR &  GTF &  SV &  LV &  SH &  TC &  BC &  ST &  SBF &  RA &  HA &  SP &  HC &  mAP \\
				\midrule
				R$^2$CNN (baseline) \cite{jiang2017r2cnn} &  80.94 &  65.67 &  35.34 &  67.44 &  59.92 &  50.91 &  55.81 &  90.67 &  66.92 &  72.39 &  55.06 &  52.23 &  55.14 &  53.35 &  48.22 &  60.67 \\
				+Pixel Attention & 81.17 & 75.23 & 36.71 & 68.14 & 62.33 & 48.22 & 55.75 & 89.57 & 78.40 & 76.61 & 54.08 & 58.32 & 63.76 & 61.94 & 54.89 & 64.34 \\
				+MDA & 84.89 & 77.07 & 38.55 & 67.88 & 61.78 & 51.87 & 56.23 & 89.82 & 75.77 & 76.30 & 53.68 & 63.25 & 63.85 & 65.05 & 53.99 & 65.33 \\
				\midrule
				+SA \cite{zhu2018seeing}+MDA & 81.27 & 76.49 & 38.16 & 69.13 & 54.03 & 46.51 & 55.03 & 89.80 & 69.92 & 75.11 & 57.06 & 58.51 & 62.70 & 59.72 & 48.20 & 62.78 \\
				+SJ \cite{zhu2018seeing}+MDA & 81.13 & 76.02 & 32.79 & 66.94 & 60.73 & 48.12 & 54.86 & 90.29 & 74.54 & 76.25 & 54.00 & 57.27 & 63.87 & 60.24 & 43.48 & 62.70 \\
				+BU \cite{zhu2018seeing} +MDA& 84.63 & 75.34 & 42.84 & 68.47 & 63.11 & 53.69 & 57.13 & 90.70 & 76.93 & 75.28 & 55.63 & 58.28 & 64.57 & 67.10 & 49.19 & 65.53 \\
				+BUS \cite{zhu2018seeing}+MDA & 87.50 & 75.60 & 42.41 & 69.48 & 62.45 & 50.89 & 56.10 & {\bf90.87} & 78.41 & 75.68 & 58.94 & 58.68 & 63.87 & 67.38 & 52.78 & 66.07 \\
				+DC \cite{zhu2018seeing}+MDA & 87.01 & 76.66 & 42.25 & 68.95 & 62.55 & 53.62 & 56.22 & 90.83 & 78.54 & 75.49 & 58.54 & 57.17 & 63.99 & 66.77 & 57.43 & 66.40 \\
				+SF+MDA & 89.65 & 79.51 & 43.86 & 67.69 & 67.41 & 55.93 & 64.86 & 90.71 & 77.77 & 84.42 & 57.67 & 61.38 & 64.29 & 66.12 & 62.04 & 68.89 \\
				\midrule
				+SF+MDA+IoU & 89.41 & 78.83 & 50.02 & 65.59 & 69.96 & 57.63 & 72.26 & 90.73 & 81.41 & 84.39 & 52.76 & 63.62 & 62.01 & 67.62 & 61.16 & 69.83 \\
				+SF +MDA+IoU+P & {\bf89.98} & {\bf80.65} & {\bf52.09} & {\bf68.36} & {\bf68.36} & {\bf60.32} & {\bf72.41} & 90.85 & {\bf87.94} & {\bf86.86} & {\bf65.02} & {\bf66.68} & {\bf66.25} & {\bf68.24} & {\bf65.21} & {\bf72.61}\\	
				\bottomrule
		\end{tabular}}
		\vspace{-10pt}
		\caption{Ablative study of each components in our proposed method on the DOTA dataset. The short names for categories are defined as: PL-Plane, BD-Baseball diamond, BR-Bridge, GTF-Ground field track, SV-Small vehicle, LV-Large vehicle, SH-Ship, TC-Tennis court, BC-Basketball court, ST-Storage tank, SBF-Soccer-ball field, RA-Roundabout, HA-Harbor, SP-Swimming pool, and HC-Helicopter.}
		\label{table:ablation}
	\end{table*}
	
	\begin{table*}
		\centering
		\resizebox{0.98\textwidth}{!}{
			\begin{tabular}{|l|c|c|c|c|c|c|c|c|c|c|c|c|c|c|c|c|}
				\toprule
				Method &  PL &  BD &  BR &  GTF &  SV &  LV &  SH &  TC &  BC &  ST &  SBF &  RA &  HA &  SP &  HC &  mAP\\
				\midrule
				\textbf{OBB} & \multicolumn{16}{|c|}{} \\
				\midrule
				FR-O \cite{xia2018dota} & 79.09 & 69.12 & 17.17 & 63.49 & 34.20 & 37.16 & 36.20 & 89.19 & 69.60 & 58.96 & 49.4 & 52.52 & 46.69 & 44.80 & 46.30 & 52.93 \\
				R-DFPN \cite{yang2018automatic} & 80.92 & 65.82 & 33.77 & 58.94 & 55.77 & 50.94 & 54.78 & 90.33 & 66.34 & 68.66 & 48.73 & 51.76 & 55.10 & 51.32 & 35.88 & 57.94 \\
				R$^2$CNN \cite{jiang2017r2cnn} & 80.94 & 65.67 & 35.34 & 67.44 & 59.92 & 50.91 & 55.81 & 90.67 & 66.92 & 72.39 & 55.06 & 52.23 & 55.14 & 53.35 & 48.22 & 60.67 \\
				RRPN \cite{ma2018arbitrary} & 88.52 & 71.20 & 31.66 & 59.30 & 51.85 & 56.19 & 57.25 & 90.81 & 72.84 & 67.38 & 56.69 & 52.84 & 53.08 & 51.94 & 53.58 & 61.01 \\
				ICN \cite{azimi2018towards} & 81.40 & 74.30 & 47.70 & 70.30 & 64.90 & 67.80 & 70.00 & 90.80 & 79.10 & 78.20 & 53.60 & 62.90 & {\bf 67.00} & 64.20 & 50.20 & 68.20 \\
				RoI-Transformer \cite{ding2018learning} & 88.64 & 78.52 & 43.44 & {\bf 75.92} & {\bf 68.81} & {\bf 73.68} & {\bf 83.59} & 90.74 & 77.27 & 81.46 & 58.39 & 53.54 & 62.83 & 58.93 & 47.67 & 69.56 \\
				SCRDet (proposed) & {\bf89.98} & {\bf80.65} & {\bf52.09} & 68.36 & 68.36 & 60.32 & 72.41 & {\bf90.85} & {\bf87.94} & {\bf86.86} & {\bf65.02} & {\bf66.68} & 66.25 & {\bf68.24} & {\bf65.21} & {\bf72.61}\\
				\midrule
				\textbf{HBB} & \multicolumn{16}{|c|}{} \\
				\midrule
				SSD \cite{fu2017dssd} & 44.74 & 11.21 & 6.22 & 6.91 & 2.00 & 10.24 & 11.34 & 15.59 & 12.56 & 17.94 & 14.73 & 4.55 & 4.55 & 0.53 & 1.01 & 10.94 \\
				YOLOv2 \cite{redmon2016you} & 76.90 & 33.87 & 22.73 & 34.88 & 38.73 & 32.02 & 52.37 & 61.65 & 48.54 & 33.91 & 29.27 & 36.83 & 36.44 & 38.26 & 11.61 & 39.20 \\
				R-FCN \cite{dai2016r} & 79.33 & 44.26 & 36.58 & 53.53 & 39.38 & 34.15 & 47.29 & 45.66 & 47.74 & 65.84 & 37.92 & 44.23 & 47.23 & 50.64 & 34.90 & 47.24 \\
				FR-H \cite{ren2017faster} & 80.32 & 77.55 & 32.86 & 68.13 & 53.66 & 52.49 & 50.04 & 90.41 & 75.05 & 59.59 & 57.00 & 49.81 & 61.69 & 56.46 & 41.85 & 60.46 \\
				FPN \cite{lin2017feature} & 88.70 & 75.10 & 52.60 & 59.20 & 69.40 & {\bf 78.80} & {\bf 84.50} & 90.60 & 81.30 & 82.60 & 52.50 & 62.10 & {\bf 76.60} & 66.30 & 60.10 & 72.00 \\
				ICN \cite{azimi2018towards} & 90.00 & 77.70 & 53.40 & {\bf 73.30} & {\bf 73.50} & 65.00 & 78.20 & 90.80 & 79.10 & 84.80 & 57.20 & 62.10 & 73.50 & 70.20 & 58.10 & 72.50 \\
				SCRDet (proposed) & {\bf 90.18} & {\bf 81.88} & {\bf 55.30} & 73.29 & 72.09 & 77.65 & 78.06 & {\bf 90.91} & {\bf 82.44} & {\bf 86.39} & {\bf 64.53} & {\bf 63.45} & 75.77 & {\bf 78.21} & {\bf 60.11} & {\bf 75.35} \\
				\bottomrule
		\end{tabular}}
		\vspace{-10pt}	
		\caption{Performance evaluation of OBB and HBB task on DOTA datasets.}
		\label{table:OBB_HBB}
	\end{table*}
	
	\begin{figure}[!tb]
		\begin{center}
			\includegraphics[width=0.87\linewidth]{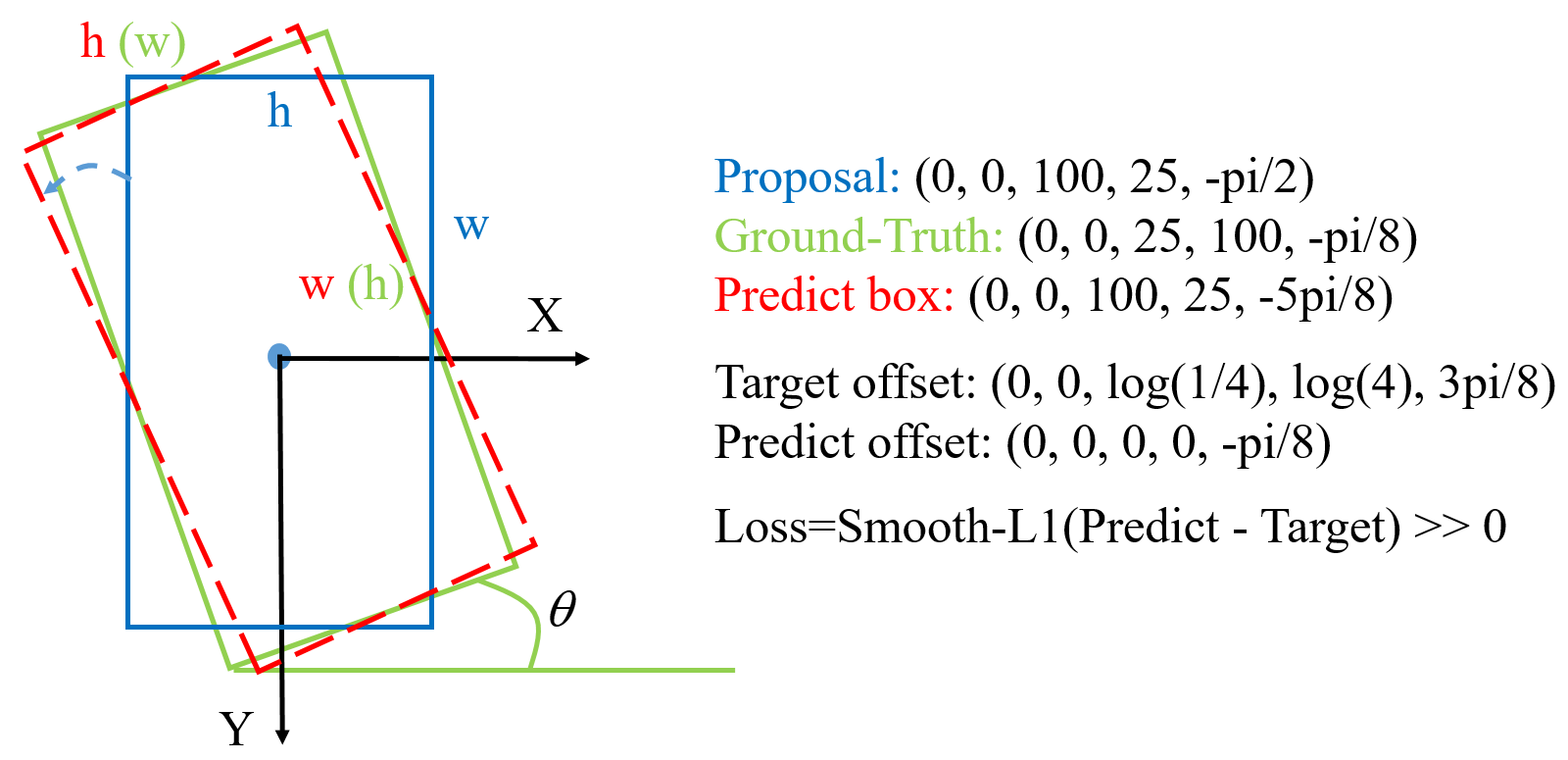}
			\vspace{-10pt}
			\caption{Boundary discontinuity of the rotation angle.}
			\label{fig:example}
		\end{center}
	\end{figure}
	
	\subsection{Loss Function}
	The multi-task loss is used which is defined as follows:
	\begin{equation}
	\begin{aligned}
	L  =& \frac{\lambda_{1}}{N}\sum_{n=1}^{N}t_{n}^{'}\sum_{j\in\{x,y,w,h,\theta\}}\frac{L_{reg}(v_{nj}^{'},v_{nj})}{|L_{reg}(v_{nj}^{'},v_{nj})|}|-\log(IoU)| \\	
	& +\frac{\lambda_{2}}{h\times w}\sum_{i}^{h}\sum_{j}^{w}L_{att}(u_{ij}^{'},u_{ij})+\frac{\lambda_{3}}{N}\sum_{n=1}^{N}L_{cls}(p_{n},t_{n})
	\label{eq:multitask_loss}
	\end{aligned}
	\end{equation}
	where $N$ indicates the number of proposals, $t_{n}$ represents the label of object, $p_{n}$ is the probability distribution of various classes calculated by Softmax function, $t_{n}^{'}$ is a binary value ($t_{n}^{'}=1$ for foreground and $t_{n}^{'}=0$ for background, no regression for background). $v_{*j}^{'}$ represents the predicted offset vectors, $v_{*j}$ represents the targets vector of ground-truth. $u_{ij}$, $u_{ij}^{'}$ represent the label and predict of mask's pixel respectively. $IoU$ denotes the overlap of the prediction box and ground-truth. The hyper-parameter $\lambda_{1}$, $\lambda_{2}$, $\lambda_{3}$  control the tradeoff. In addition, the classification loss $L_{cls}$ is Softmax cross-entropy. The regression loss $L_{reg}$ is smooth L1 loss as defined in \cite{girshick2015fast}, and the attention loss $L_{att}$ is pixel-wise Softmax cross-entropy.
	
\begin{figure}[tb!]
		\centering
		\begin{subfigure}{.23\textwidth}
			\centering
			\includegraphics[width=.9\linewidth]{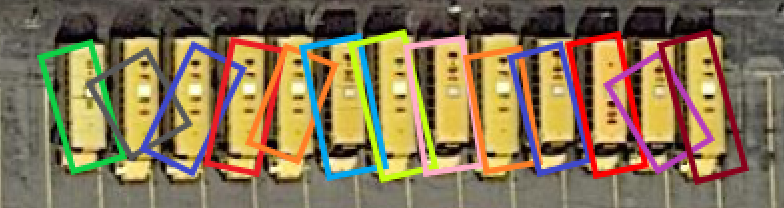}
			\caption{Smooth L1 loss}
			\label{fig:bad_case}
		\end{subfigure}
		\vspace{-3pt}
		\begin{subfigure}{.23\textwidth}
			\centering
			\includegraphics[width=.9\linewidth]{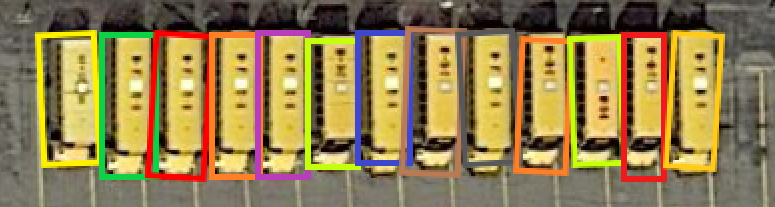}
			\caption{IoU-smooth L1 loss}
			\label{fig:good_case}
		\end{subfigure}
		\vspace{-10pt}
		\caption{Comparison of detection results by two losses.}
		\label{fig:cases}
	\end{figure}

	In particular, there exists the boundary problem for the rotation angle, as shown in Fig.~\ref{fig:example}. It shows that an ideal form of regression (the blue box rotates counterclockwise to the red box), but the loss of this situation is very large due to the periodicity of the angle. Therefore, the model has to be regressed in other complex forms (such as the blue box rotating clockwise while scaling $w$ and $h$), increasing the difficulty of regression, as shown in Fig.~\ref{fig:bad_case}. To better solve this problem, we introduce the IoU constant factor $\frac{|-\log(IoU)|}{|L_{reg}(v_{j}^{'},v_{j})|}$ in the traditional smooth L1 loss, as shown in Eq. \ref{eq:multitask_loss}. It can be seen that in the boundary case, the loss function is approximately equal to $|-\log(IoU)|\approx0$, eliminating the sudden increase in loss, as shown in Fig.~\ref{fig:good_case}. The new regression loss can be divided into two parts, $\frac{L_{reg}(v_{j}^{'},v_{j})}{|L_{reg}(v_{j}^{'},v_{j})|}$ determines the direction of gradient propagation, and $|-\log(IoU)|$ for the magnitude of  gradient. In addition, using IoU to optimize location accuracy is consistent with IoU-dominated metric, which is more straightforward and effective than coordinate regression.
	
	\section{Experiments}
	Tests are implemented by TensorFlow \cite{abadi2016tensorflow} on a server with Nvidia Geforce GTX 1080 GPU and 8G memory. We perform experiments on both aerial benchmarks and natural images to verify the generality of our techniques. Note our techniques are orthogonal to specific network backbone. In experiments, we use Resnet-101 as backbone for remote sensing benchmarks, and FPN and R$^2$CNN for COCO, VOC2007 and ICDAR2015 respectively.
	

	\subsection{Experiments on Aerial Images}
	\subsubsection{Datasets and Protocls}
	The benchmark DOTA \cite{xia2018dota} is for object detection in aerial images. It contains 2,806 aerial images from different sensors and platforms. The image size ranges from around $ 800 \times 800 $ to $ 4,000 \times 4,000 $ pixels and contains objects exhibiting a wide variety of scales, orientations, and shapes. These images are then annotated by experts using 15 common object categories. The fully annotated DOTA benchmark contains 188,282 instances, each of which is labeled by an arbitrary quadrilateral. There are two detection tasks for DOTA: horizontal bounding boxes (HBB) and oriented bounding boxes (OBB). Half of the original images are randomly selected as the training set, 1/6 as the validation set, and 1/3 as the testing set. We divide the images into $ 800 \times 800 $ subimages with an overlap of 200 pixels.
	
	The public benchmark NWPU VHR-10 \cite{cheng2016learning} contains 10-class geospatial object for detection. This dataset contains 800 very-high-resolution (VHR) remote sensing images that are cropped from Google Earth and Vaihingen dataset and then manually annotated by experts.
	
	We use the pretrained ResNet-101 model for initialization. For DOTA, the model is trained by 300k iterations in total, and the learning rate changes during the 100k and 200k iterations from 3e-4 to 3e-6. For NWPU VHR-10, the split ratios of the training dataset, validation dataset, and test dataset are 60\%, 20\%, and 20\%, respectively. The model is trained by totally 20k iterations with the same learning rate as for DOTA. Besides, weight decay and momentum are 0.0001 and 0.9, respectively. We employ MomentumOptimizer as optimizer and no data augmentation is performed except random image flip during training. 

	For parameter setting, the expected anchor stride $S$ as discussed in Sec. \ref{subsec:sfn} is set to 6, and we set the base anchor size to 256, and the anchor scales setting from $2^{-4}$ to $2^{1}$. Since the multi-categories objects in DOTA and NWPU VHR-10 have different shapes, we set anchor ratios to [1/1,1/2,1/3,1/4,1/5,1/6,1/7,1/9]. These settings ensure that each ground-truth can be assigned with positive samples. When $IoU>0.7$, the anchor is assigned as a positive sample, and as a negative sample if $IoU<0.3$. Besides, due to the sensitivity between angle and IoU in the large aspect ratio rectangle, the two thresholds in the second stage are all set to 0.4, respectively. For training, the mini-batch size in two stages is 512. The hyperparameters in Eq.~\ref{eq:multitask_loss} are set to $\lambda_{1}=4$, $\lambda_{2}=1$, $\lambda_{3}=2$.
	
	\begin{figure*}[t]
		\centering
		\begin{subfigure}{.15\textwidth}
			\centering
			\includegraphics[width=.98\linewidth, height=2.4cm]{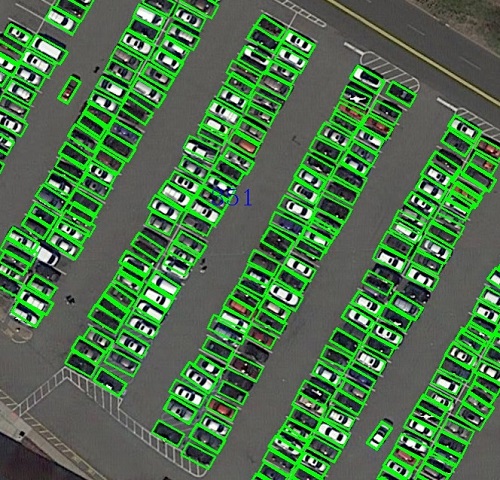}
			\caption{SV}
			\label{fig:dota_a}
		\end{subfigure}%
		\begin{subfigure}{.15\textwidth}
			\centering
			\includegraphics[width=.98\linewidth, height=2.4cm]{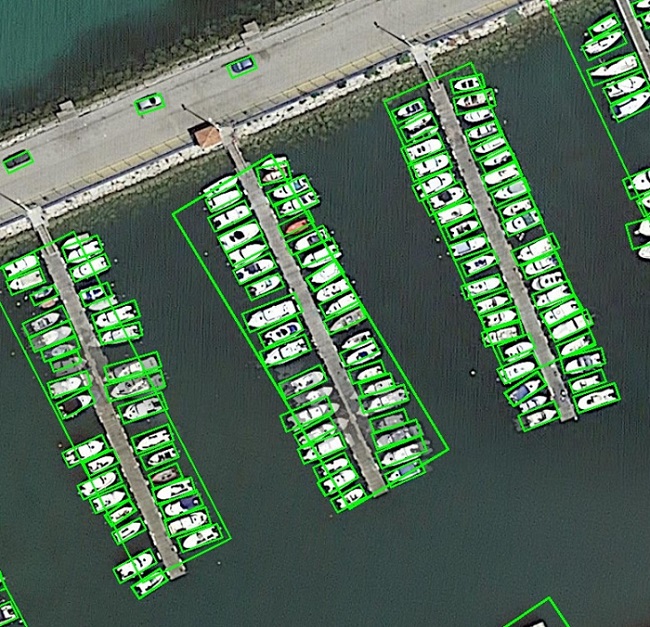}
			\caption{SH and HA}
			\label{fig:dota_b}
		\end{subfigure}
		\begin{subfigure}{.15\textwidth}
			\centering
			\includegraphics[width=.98\linewidth, height=2.4cm]{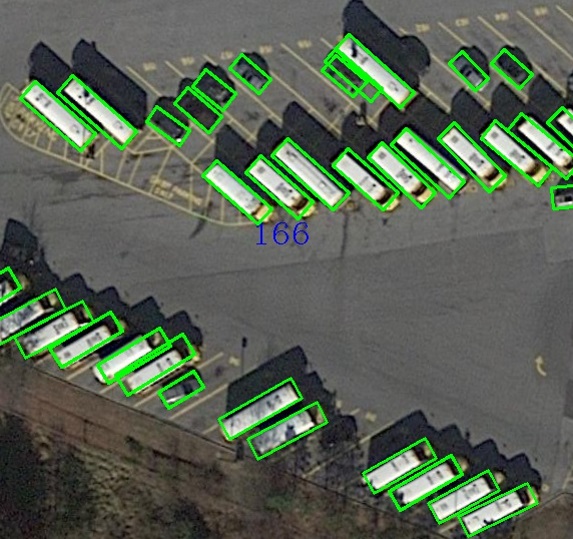}
			\caption{LV}
			\label{fig:dota_c}
		\end{subfigure}%
		\begin{subfigure}{.15\textwidth}
			\centering
			\includegraphics[width=.98\linewidth, height=2.4cm]{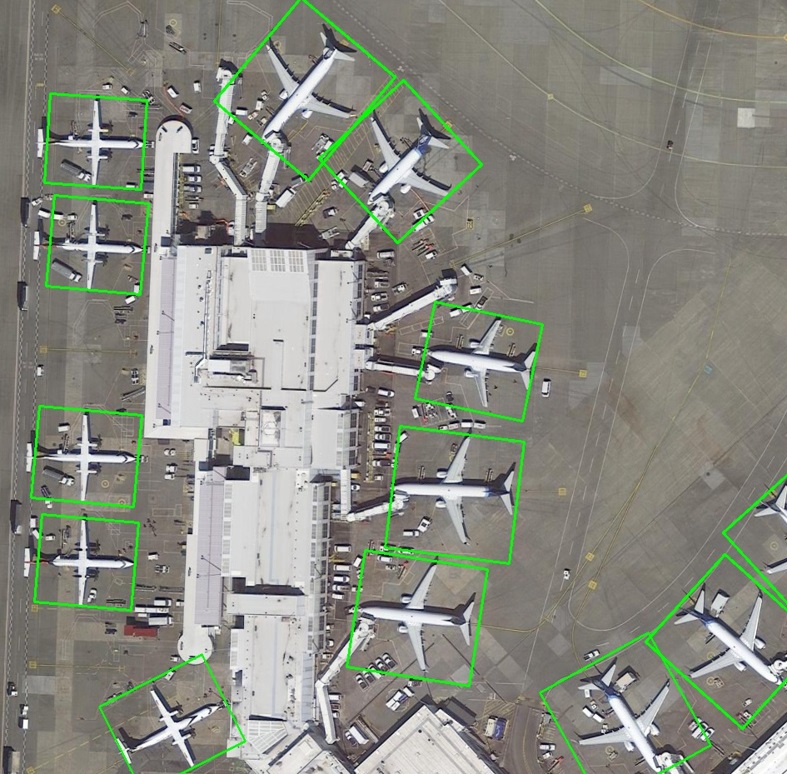}
			\caption{PL}
			\label{fig:dota_d}
		\end{subfigure}
		\begin{subfigure}{.15\textwidth}
			\centering
			\includegraphics[width=.99\linewidth, height=2.4cm]{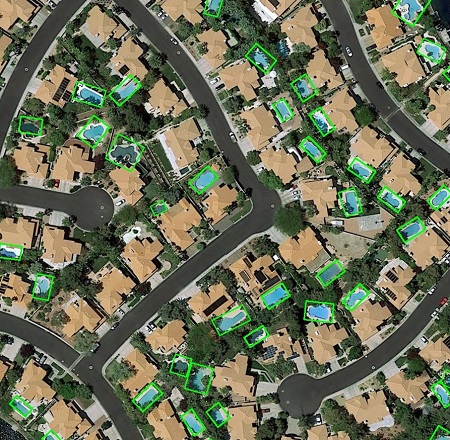}
			\caption{SP}
			\label{fig:dota_e}
		\end{subfigure}%
		\begin{subfigure}{.15\textwidth}
			\centering
			\includegraphics[width=.97\linewidth, height=2.4cm]{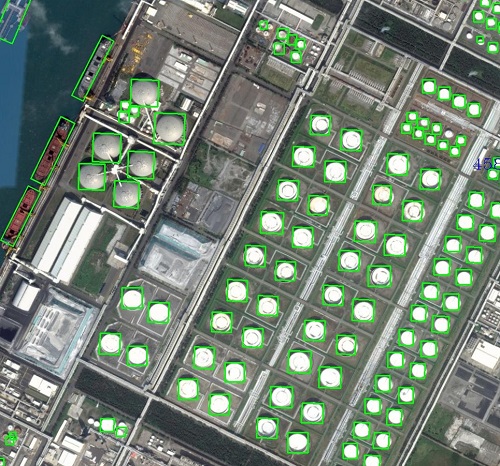}
			\caption{ST}
			\label{fig:dota_f}
		\end{subfigure}\\
		\begin{subfigure}{.15\textwidth}
			\centering
			\includegraphics[width=.99\linewidth, height=2.4cm]{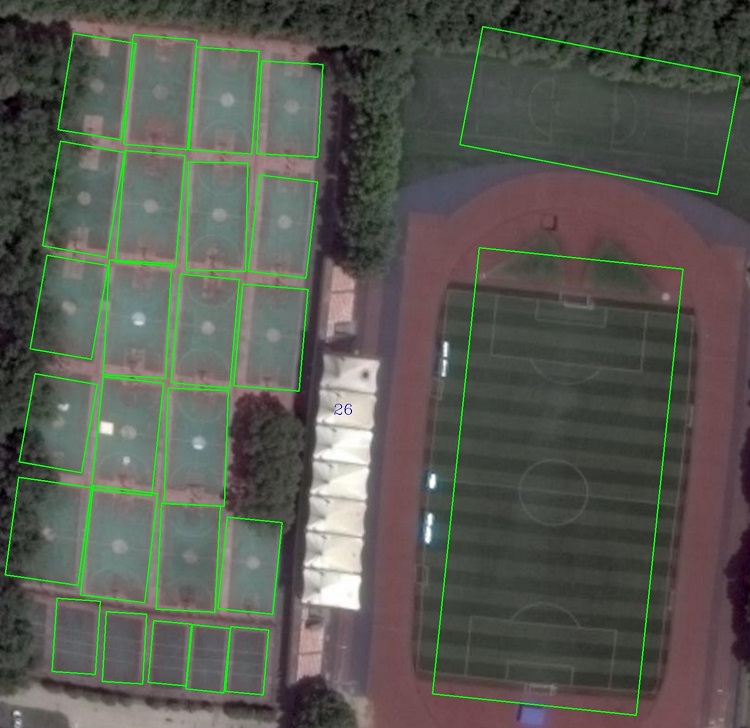}
			\caption{SBF}
			\label{fig:dota_g}
		\end{subfigure}%
		\begin{subfigure}{.15\textwidth}
			\centering
			\includegraphics[width=.97\linewidth, height=2.4cm]{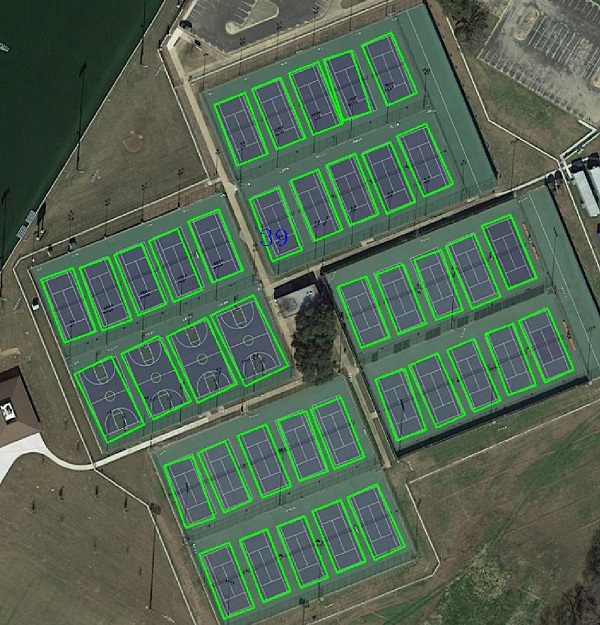}
			\caption{TC and BC}
			\label{fig:dota_h}
		\end{subfigure}
		\begin{subfigure}{.15\textwidth}
			\centering
			\includegraphics[width=.99\linewidth, height=2.4cm]{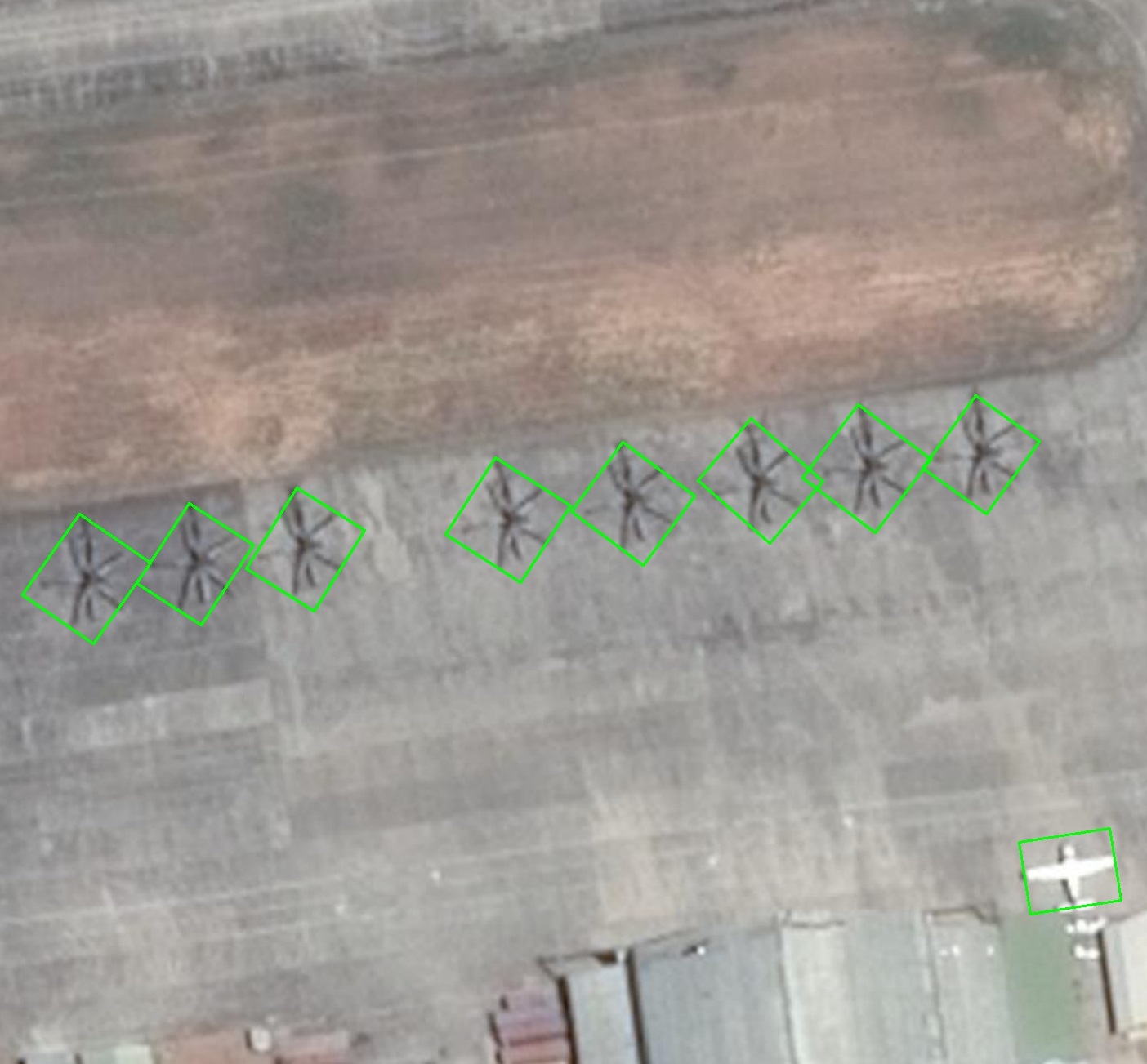}
			\caption{HC}
			\label{fig:dota_i}
		\end{subfigure}%
		\begin{subfigure}{.15\textwidth}
			\centering
			\includegraphics[width=.97\linewidth, height=2.4cm]{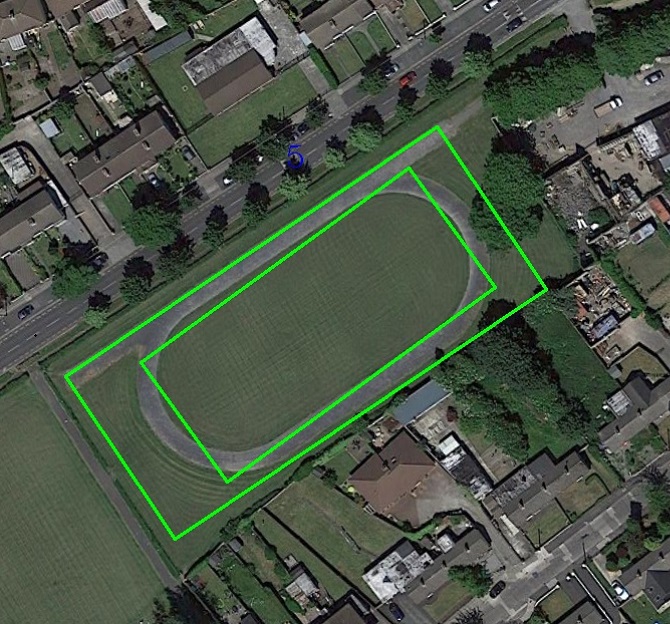}
			\caption{GTF}
			\label{fig:dota_j}
		\end{subfigure}
		\begin{subfigure}{.15\textwidth}
			\centering
			\includegraphics[width=.99\linewidth, height=2.4cm]{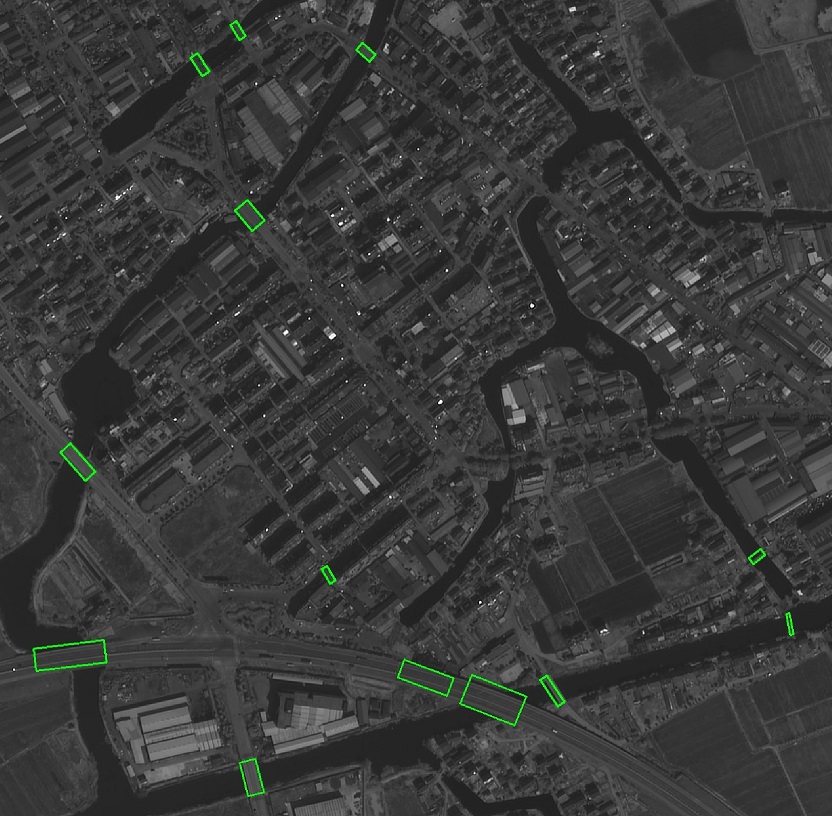}
			\caption{BR}
			\label{fig:dota_k}
		\end{subfigure}%
		\begin{subfigure}{.15\textwidth}
			\centering
			\includegraphics[width=.97\linewidth, height=2.4cm]{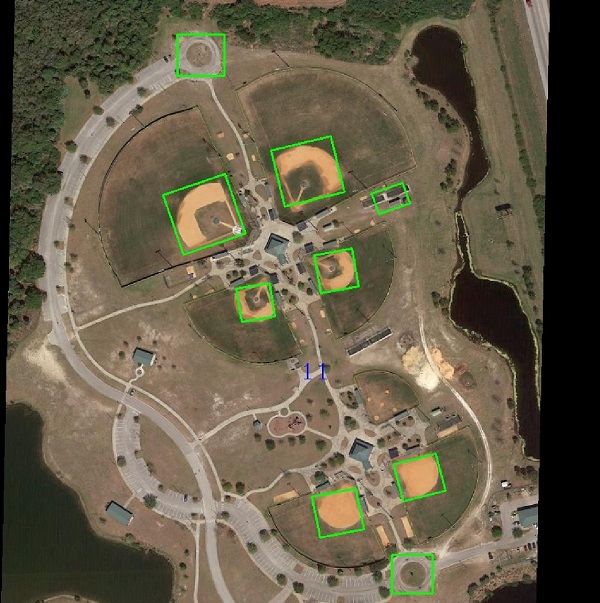}
			\caption{BD and RA}
			\label{fig:dota_l}
		\end{subfigure}
		\vspace{-10pt}
		\caption{Examples on DOTA. Our method performs better on those with small size, in arbitrary direction, and high density.}	\label{fig:dota}
	\end{figure*}

	\subsubsection{Ablation Study}
	{\bf \quad Baseline setup.} We choose Faster-RCNN-based R$^2$CNN \cite{jiang2017r2cnn} as the baseline for ablation study, but not limited to this method. For fairness, all experimental data and parameter settings are strictly consistent. We use mean average precision (mAP) as a measure of performance. The results of DOTA reported here are obtained by submitting our predictions to the official DOTA evaluation server\footnote{https://captain-whu.github.io/DOTA/}.
	
	{\bf Effect of MDA-Net.} As discussed in Sec. \ref{subsec:mdan}, the attention structure is beneficial to suppress the influence of noise and highlight the object information. It also can be evidenced in Table \ref{table:ablation} that the detection results of most objects have been improved to varying degrees after adding the pixel attention network, and the total mAP increase by 3.67\%. MDA-Net further improves the detection accuracy of large aspect ratio targets such as bridge, large vehicle, ship, harbor and so on. Compared to pixel attention, MDA-Net increases mAP by about 1\% to 65.33\%. Table \ref{table:MDA-Net} shows that supervised learning is the main contribution of MDA-Net rather than computation.
	
	{\bf Effect of SF-Net.} Reducing the stride size of the anchor and the feature fusion are effective means to improve the detection for small objects. In Table \ref{table:ablation} we also study on the techniques presented in \cite{zhu2018seeing}. Both shifted anchors (SA) and shift jittering (SJ) follow the idea of using a single feature point to regress the bounding boxes of multiple sub-areas. Experiments show that these two strategies can hardly contribute to the accuracy in accordance with the observation in the original paper. Enlarged feature maps is a good strategy to reduce $S_{A}$, including bilinear upsampling (BU), bilinear upsampling with skip connection (BUS), dilated convolution (DC). Although these methods take into account the importance of sampling for small object detection and their detection performance have been improved to varying degrees, the  $S_{A}$ settings are still inflexible and cannot achieve the best sampling results. SF-Net effectively models the feature fusion and the flexibility of the $S_{A}$ setting, and it achieves the best performance of 68.89\%, especially benefited from the improvement of small object such as vehicle, ship and storage tank.
	
	{\bf Effect of IoU-Smooth L1 Loss.} IoU-Smooth L1 Loss  eliminates the boundary effects of the angle, making it easier for the model to regress to the objects coordinates. This new loss improves the detection accuracy to 69.83\%.
	
	{\bf Effect of image pyramid.} Image pyramid based training and test is an effective means to improve performance. The method ICN \cite{azimi2018towards} uses the image cascade network structure, which is similar to the idea of image pyramid. Here we randomly scale the original image to [$600\times600$, $800\times800$, $1,000\times1,000$, $1,200\times1,200$] and send it to the network for training. For testing, each image is tested at four scales and combined by R-NMS. As shown in Table \ref{table:ablation}, image pyramid can notably improve the detection efficiency and achieves 72.61\% mAP. The detection results for each class on DOTA are shown in Fig.~\ref{fig:dota}.
	
	\begin{table}[!tb]
		\centering
		\resizebox{0.32\textwidth}{!}{
			\begin{tabular}{ll}
				\toprule
				Method & mAP\\
				\midrule
				R-P-Faster R-CNN \cite{han2017efficient} & 76.50 \\
				SSD512 \cite{liu2016ssd} & 78.40 \\
				DSSD321 \cite{fu2017dssd} & 78.80 \\
				DSOD300 \cite{shen2017dsod} & 79.80 \\
				Deformable R-FCN \cite{xu2017deformable} & 79.10 \\
				Deformable Faster R-CNN \cite{ren2018deformable} & 84.40 \\
				RICADet \cite{li2018rotation} & 87.12 \\
				RDAS512 \cite{chen2018geospatial} & 89.50 \\
				Multi-Scale CNN \cite{guo2018geospatial} & 89.60 \\
				SCRDet (proposed) & {\bf 91.75} \\
				\bottomrule
		\end{tabular}}
		\vspace{-10pt}
		\caption{Performance for HBB task on NWPU VHR-10.}
		\label{table:HBB_NWPU}
	\end{table}
	
	\begin{table}
		\begin{center}
			\resizebox{0.45\textwidth}{5mm}{
				\begin{tabular}{|l|c|c|c|c|}
					\hline
					dataset train/test & baseline & MDA-Net &  MDA-Net$^\dag$ & baseline$^\dag$ \\
					\hline
					DOTA trainval/test & 60.67\% (R$^2$CNN) & {\bf 65.33\%} & 61.23\% & 65.08\% \\
					\hline
					VOC 07+12/07 & 80.39\% (FPN$^*$) & {\bf 82.27\%} & 80.53\% & 82.11\% \\
					\hline
			\end{tabular}}
		\end{center}
		\vspace{-18pt}
		\caption{MDA-Net$^\dag$ means MDA-Net without supervised learning. baseline$^\dag$ means baseline with supervision.} 
		\label{table:MDA-Net}
	\end{table}
	
	\subsubsection{Peer Methods Comparison}

	\begin{figure}[!tb]
		\centering
		\begin{subfigure}{.23\textwidth}
			\centering
			\includegraphics[width=4cm, height=2.2cm]{figure/COCO_val2014_000000245915.jpg}
			\caption{}
			\label{fig:COCO_val2014_000000245915}
		\end{subfigure}
		\begin{subfigure}{.23\textwidth}
			\centering
			\includegraphics[width=4cm, height=2.2cm]{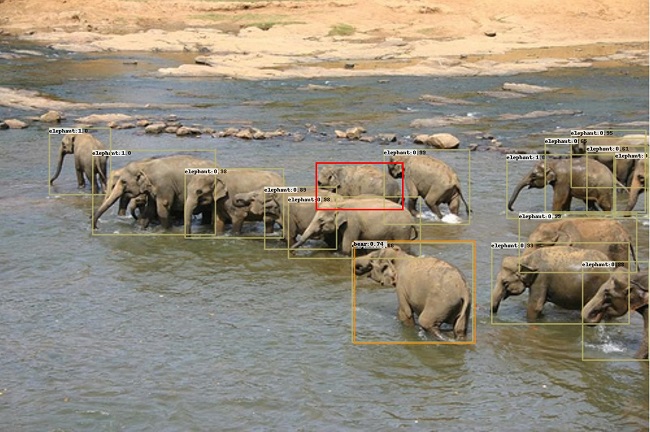}
			\caption{}
			\label{fig:COCO_val2014_000000245915_bs}
		\end{subfigure} \\
		\vspace{-10pt}
		\begin{subfigure}{.23\textwidth}
			\centering
			\includegraphics[width=4cm, height=2.2cm]{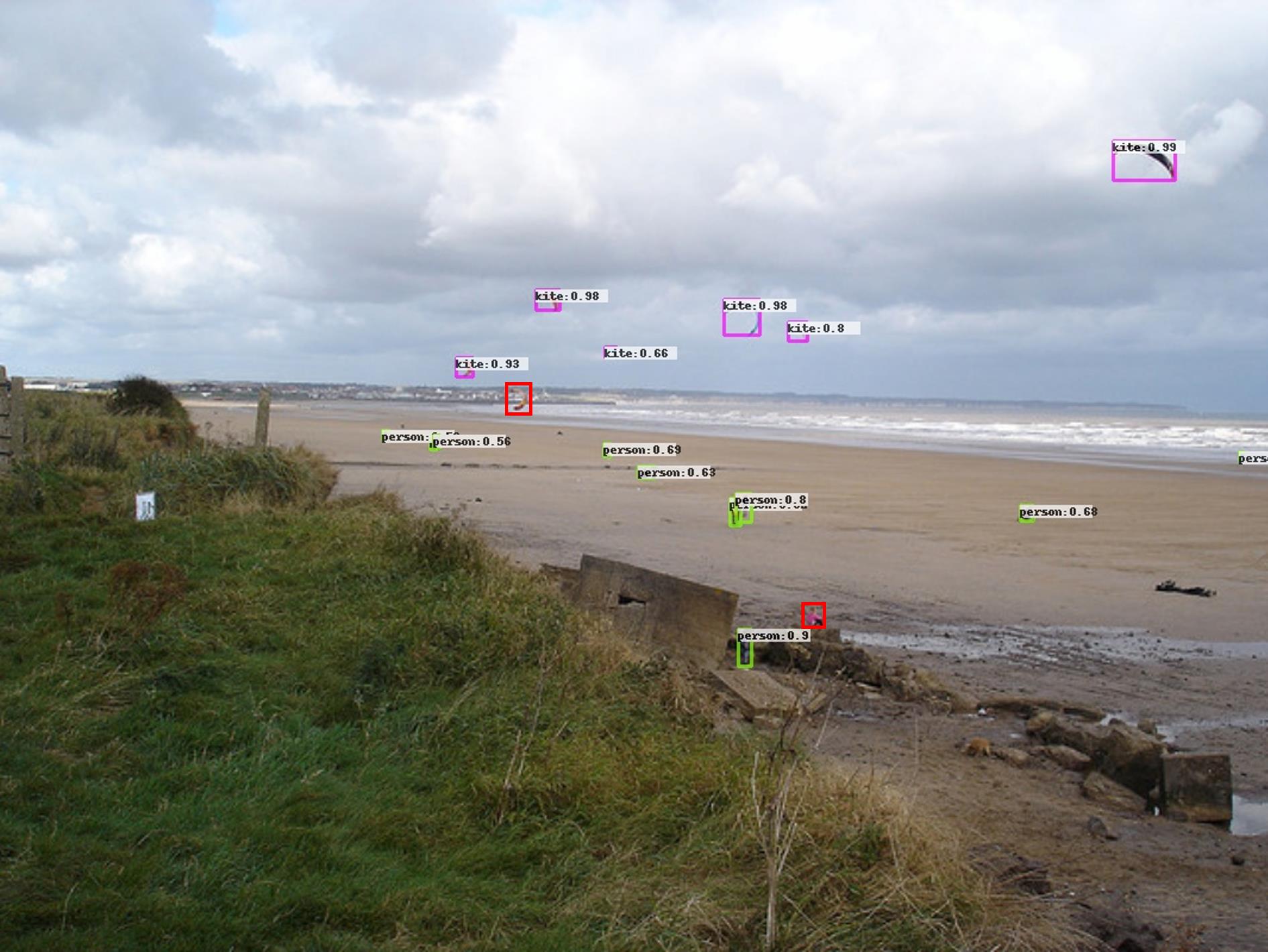}
			\caption*{FPN$^*$+MDA-Net}
			\label{fig:COCO_val2014_000000224664}
		\end{subfigure}
		\begin{subfigure}{.23\textwidth}
			\centering
			\includegraphics[width=4cm, height=2.2cm]{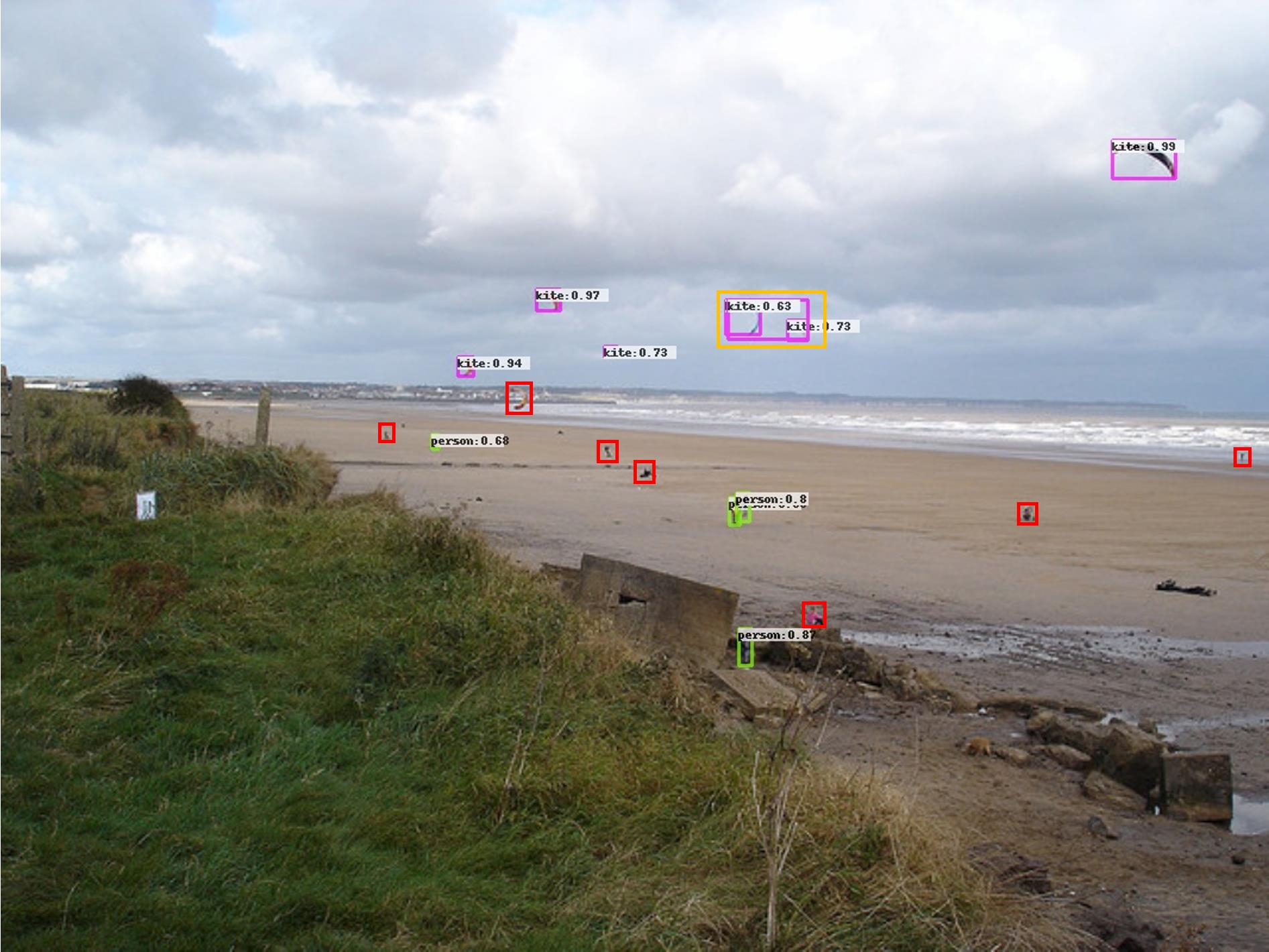}
			\caption*{FPN$^*$}
			\label{fig:COCO_val2014_000000224664_bs}
		\end{subfigure}
		\vspace{-10pt}
		\caption{Detection results of COCO. The first column is the result of FPN$^*$+MDA-Net and the second column is FPN$^*$. The red boxes represent missed objects and the orange boxes represent false alarm.}
		\label{fig:coco_det}
	\end{figure}
	
	\begin{figure}[!tb]
		\vspace{-10pt}
		\centering
		\begin{subfigure}{.23\textwidth}
			\centering
			\includegraphics[width=4cm, height=2.2cm]{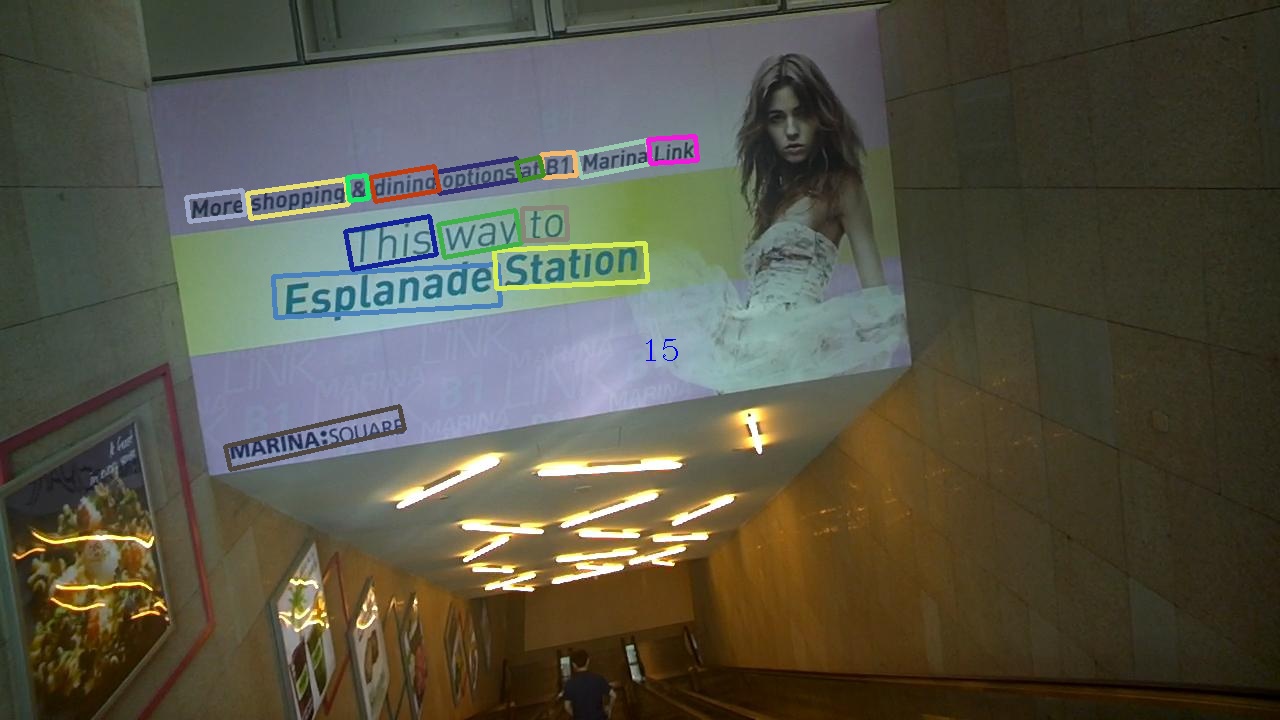}
			\caption{}
			\label{fig:img_108_R2CNN++}
		\end{subfigure}
		\begin{subfigure}{.23\textwidth}
			\centering
			\includegraphics[width=4cm, height=2.2cm]{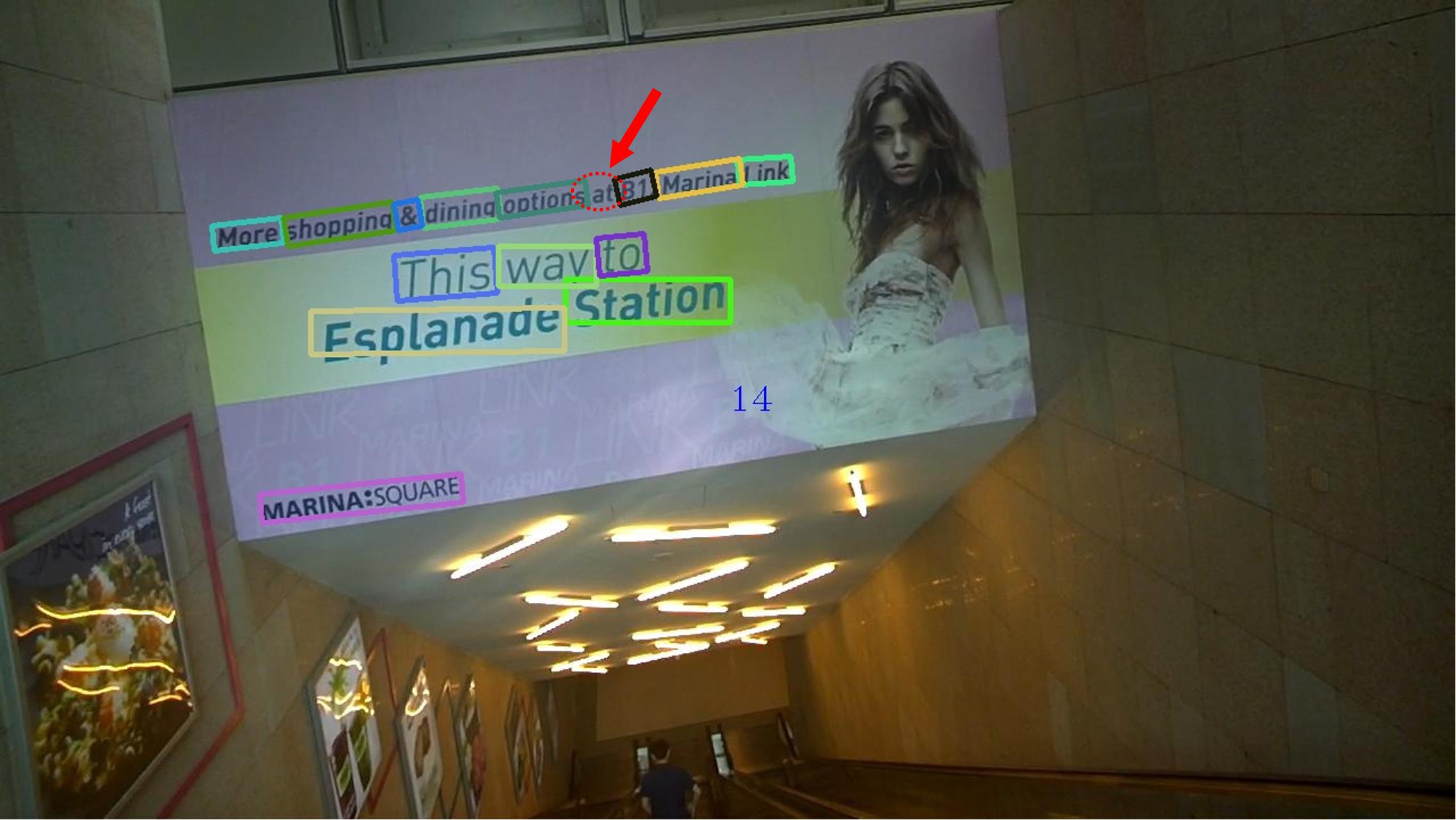}
			\caption{}
			\label{fig:img_108_R2CNN_}
		\end{subfigure} \\
		\vspace{-10pt}
		\begin{subfigure}{.23\textwidth}
			\centering
			\includegraphics[width=4cm, height=2.2cm]{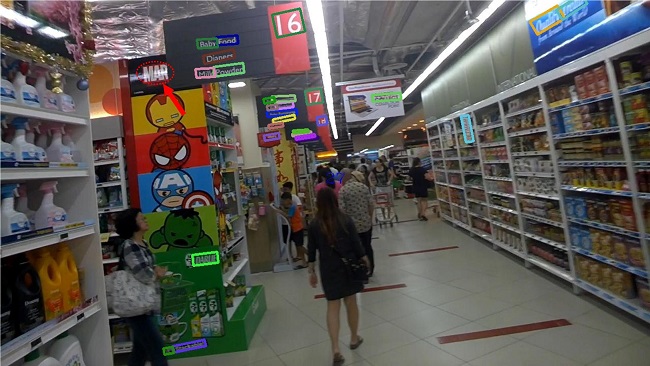}
			\caption*{SCRDet-R$^2$CNN}
			\label{fig:img_51_R2CNN++_}
		\end{subfigure}
		\begin{subfigure}{.23\textwidth}
			\centering
			\includegraphics[width=4cm, height=2.2cm]{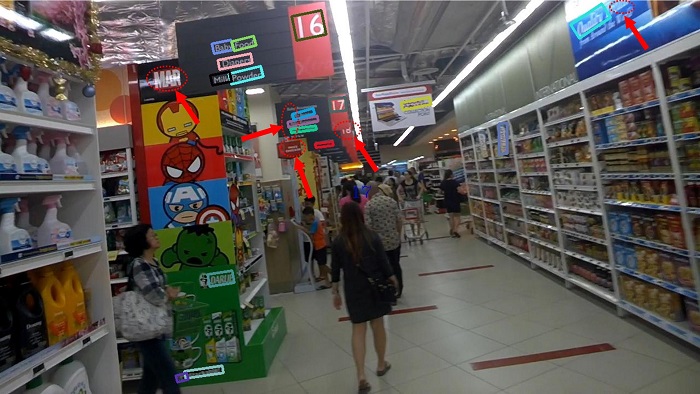}
			\caption*{R$^2$CNN-4$^*$}
			\label{fig:img_51_R2CNN_}
		\end{subfigure}
		\vspace{-10pt}
		\caption{Detection results of COCO and ICDAR2015. The first column is the result of R$^2$CNN-4$^*$ equipped with our techniques (SCRDet-R$^2$CNN) and the second column is vanilla R$^2$CNN-4$^*$. Red arrows denote missed objects.}
		\label{fig:icdar2015}
	\end{figure}
	
	{\bf \quad OBB Task.} Besides the official baseline given by DOTA, we also compare with RRPN \cite{ma2018arbitrary}, R$^2$CNN \cite{jiang2017r2cnn}, R-DFPN \cite{yang2018automatic}, ICN \cite{azimi2018towards} and RoI-Transformer~\cite{ding2018learning}, which are all applicable to multi-category rotation object detection. Table \ref{table:OBB_HBB} shows the performance of these methods. The excellent performance of RoI-Transformer, ICN and SCRDet in small object detection is attributed to feature fusion. SCRDet considers the expansion of the receptive field and the attenuation of noise in the fusion, so it is better than ICN and RoI-Transformer for large objects. Our method ranks first among existing published results, reaching 72.61\% mAP.
	
	{\bf HBB Task.} We use DOTA and NWPU VHR-10 to validate our proposed approach and shield the angle parameter in the code. Table \ref{table:OBB_HBB} and Table \ref{table:HBB_NWPU} show the performance on the two datasets, respectively. We also get the first place among existing methods in literature on DOTA, at 75.35\% or so. For the NWPU VHR-10 dataset, we compare it with nine methods and achieve the best detection performance, at 91.75\%. Our approach achieves the best detection accuracy on more than half of the categories. 
	
	\subsection{Experiments on Natural Images}\label{sec:ee}
	To verify the universality of our model, we further validate the proposed techniques on generic datasets and general-purpose detection networks FPN \cite{lin2017feature} and R$^2$CNN \cite{jiang2017r2cnn}. We choose COCO \cite{lin2014microsoft} and VOC2007 \cite{everingham2010pascal} datasets as they contain many small objects. We also use ICDAR2015 \cite{karatzas2015icdar} because there are rotated texts for scene text detection.
	
	By Table \ref{table:experiments}, FPN$^*$ with MDA-Net can increase by 0.7\% and 2.22\% on COCO \cite{lin2014microsoft} and VOC2007 \cite{everingham2010pascal} datasets, respectively. As shown in Fig.~\ref{fig:coco_det}, the MDA-Net has good performance in both dense and small objects detection. IoU-Smooth loss does not bring high improvement to horizontal region detection, hence this also reflects its pertinence to rotation detection boundary problem.
	
	For ICDAR2015, R$^2$CNN-4 achieves 74.36\% in single scale according to \cite{jiang2017r2cnn}. As it is not open sourced, we reimplement it and term our version as R$^2$CNN-4$^*$ according to the definition of the rotation box in the paper without multiple pooled sizes structure, and our version can achieve the mAP of 77.23\%. Then, we equip R$^2$CNN-4$^*$ with our proposed techniques and term it SCRDet-R$^2$CNN. It achieves the highest performance 80.08\% in single scale. Once again, the validity of the structure proposed in this paper is proved. According to Fig.~\ref{fig:icdar2015}, SCRDet-R$^2$CNN, achieves a notably better recall for dense objects detection. 
	\begin{table}[]
		\centering
		\resizebox{0.46\textwidth}{!}{
			\begin{tabular}{|l|c|c|c|}
				\hline
				Dataset &  Model &  Backbone  &  mAP/F1 \\
				\hline
				\multirow{3}{*}{COCO}
				& FPN$^*$ & Res50  & 36.1 \\ \cline{2-4}
				& FPN$^*$+IoU-Smooth & Res50  & 36.2 \\ \cline{2-4}
				& FPN$^*$+MDA-Net & Res50  & {\bf36.8} \\ \cline{2-4}
				\hline
				\multirow{2}{*}{VOC2007}
				& FPN$^*$ & Res101  & 76.14 \\ \cline{2-4}
				& FPN$^*$+MDA-Net & Res101  & {\bf78.36} \\
				\hline
				\multirow{2}{*}{ICDAR2015}
				& R$^2$CNN-4$^*$ & Res101 & 77.23 \\ \cline{2-4}
				& SCRDet-R$^2$CNN & Res101  & {\bf80.08} \\
				\hline
		\end{tabular}}
		\vspace{-10pt}
		\caption{Effectiveness of the proposed structure on generic datasets. Notation $^*$ indicates our own implementation. For VOC 2007, all methods are trained on VOC2007 trainval sets and tested on VOC 2007 test set. For COCO, all the results are obtained on the $minival$ set. For ICDAR2015, results are obtained by submitting it to the official website.}
		\label{table:experiments}
	\end{table}
	
	\section{Conclusion}
	We have presented an end-to-end multi-category detector designated for objects in arbitrary rotations, which are common in aerial image. Considering the factors of feature fusion and anchor sampling, a sampling fusion network with smaller $S_{A}$ is added. Meanwhile, the algorithm weakens the influence of noise and highlights the object information through a supervised multi-dimensional attention network. Moreover, we implement rotation detection to preserve orientation information and solve intensive problems. Our approach achieves state-of-the-art performance on two public remote sensing datasets: DOTA and NWPU VHR-10. Finally, we have further validated our structure on nature datasets such as COCO, VOC2007 and ICDAR2015.
	
	{\small
		\bibliographystyle{ieee_fullname}
		\bibliography{egbib}
	}
	
\end{document}